\pdfoutput=1

\documentclass[11pt]{article}

\usepackage[preprint]{acl}

\usepackage{times}
\usepackage{latexsym}
\usepackage{soul}

\usepackage[T1]{fontenc}

\usepackage[utf8]{inputenc}

\usepackage{microtype}

\usepackage{inconsolata}

\usepackage{graphicx}

\usepackage{adjustbox}
\usepackage{amsmath}
\usepackage{amssymb}
\usepackage[scaled=0.8]{beramono}
\usepackage{booktabs}
\usepackage{graphicx}
\usepackage{multicol}
\usepackage{multirow}
\usepackage{todonotes}
\usepackage{url}
\usepackage{xcolor}
\usepackage{xspace}
\usepackage{enumitem}
\usepackage{tcolorbox}

\newcommand{\dataset}{\textsc{ClaimCheck}\xspace}

\definecolor{original}{RGB}{108, 86, 140}
\definecolor{revised}{RGB}{129, 140, 116}
\title{\textsc{ClaimCheck}: How Grounded are LLM Critiques of Scientific Papers?}

\author{
\bf Jiefu Ou$^*$, William Walden$^*$,\\ 
\bf Kate Sanders, Zhengping Jiang, Kaiser Sun, Jeffrey Cheng, \\ 
\bf William Jurayj, Miriam Wanner, Shaobo Liang, Candice Morgan, \\ 
\bf Seunghoon Han, Weiqi Wang, Chandler May, Hannah Recknor, \\ 
\bf Daniel Khashabi, Benjamin Van Durme \\
Johns Hopkins University \\
\texttt{jou6@jhu.edu}}

\begin{document}
\maketitle
\def\thefootnote{*}\footnotetext{These authors contributed equally to this work}\def\thefootnote{\arabic{footnote}}
\begin{abstract}
A core part of scientific peer review involves providing expert critiques that directly assess the scientific claims a paper makes. While it is now possible to automatically generate plausible (if generic) reviews, ensuring that these reviews are \emph{sound} and \emph{grounded} in the papers' claims remains challenging.
To facilitate LLM benchmarking on these challenges, we introduce \dataset, an annotated dataset of NeurIPS 2023 and 2024 submissions and reviews mined from OpenReview. \dataset is richly annotated by ML experts for weakness statements in the reviews and the paper claims that they dispute, as well as fine-grained labels of the validity, objectivity, and type of the identified weaknesses.
We benchmark several LLMs on three claim-centric tasks supported by \dataset, requiring models to (1) associate weaknesses with the claims they dispute, (2) predict fine-grained labels for weaknesses and rewrite the weaknesses to enhance their specificity, and (3) verify a paper's claims with grounded reasoning. Our experiments reveal that cutting-edge LLMs, while capable of predicting weakness labels in (2), continue to underperform relative to human experts on all other tasks.\footnote{\url{https://github.com/JHU-CLSP/CLAIMCHECK}}
\end{abstract}

\section{Introduction}
\label{sec:introduction}
Scientific peer review demands expert critique of the \emph{claims} that a paper makes---about results, theorems, approaches, novelty, etc. It is thus paramount that these critiques be \emph{grounded} in the actual assertions made by a papers' authors, a principle reflected in the guidelines for numerous top-tier AI conferences (\autoref{tab:reviewer-guideline}). Curiously, however, the literature on automated peer review has paid little attention to the problem of ensuring that reviews adhere to this principle (see \S\ref{sec:background}). As LLMs encroach ever more into knowledge-intensive work---including peer review---adequately addressing the challenge of producing grounded generations is paramount.

\begin{figure}[t]
    \centering
    \includegraphics[width=\columnwidth]{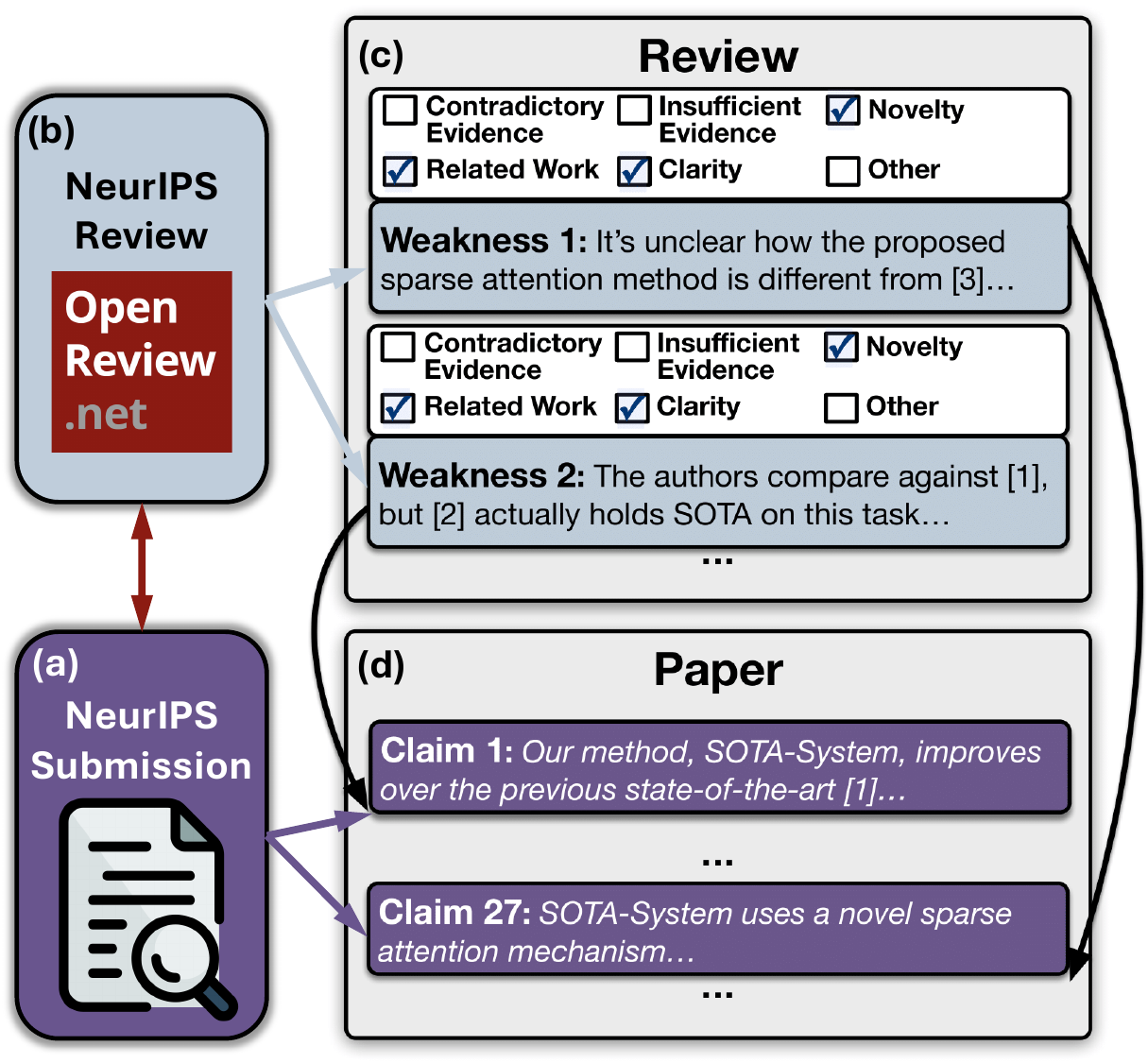}
    \caption{\dataset is sourced from rejected NeurIPS submissions (a) and their corresponding reviews from OpenReview (b).  Annotations identify \emph{claim-related weaknesses} 
    in reviews (blue arrows to Weakness 1 \& 2) 
    and provide fine-grained labels (c).  Weaknesses are grounded to specific \emph{target claims} (Claim 27 \& 1, respectively) that they dispute (black arrows from (c) to (d)). These \emph{target claims} are identified from a set of claims extracted from the original paper (purple arrows to from (a) to (d)). Grounding weaknesses in a paper's claims is essential in peer review.
    }
    \vspace{-5mm}
    \label{fig:fig-1}
\end{figure}

However, collecting authentic data for grounded peer review is challenging. Existing work tends to simplify this challenge by narrowing either the scope of the claims or the evidence pool \citep{wadden-etal-2020-fact, lu-etal-2023-scitab}, making systems trained on such data difficult to deploy in real-world peer review scenarios. Further, these approaches tend to employ binary claim factuality judgments, limiting their applicability to the common case in which a claim is \textit{flawed} but not entirely \textit{false}~\cite{estornell2020deception, venkat2022liarx}.

To address these challenges, we introduce  \dataset, a novel resource for \emph{automatic, claim-grounded peer review}. \dataset is a high-quality, multimodal collection of rejected NeurIPS submissions and their reviews, annotated by ML experts for rich information about reviewer-identified, \emph{claim-related} weaknesses, with links to the paper claims they dispute (\autoref{fig:fig-1}). To our knowledge, \dataset is the first resource that supports both scientific claim verification and claim-grounded peer review. Claims are sourced directly from papers' full texts, rather than synthetically constructed \citep{thorne-etal-2018-fever, thorne-etal-2018-fact}. Further, as these claims are rarely \emph{entirely true} or \emph{entirely false}, we develop an informative, multi-label ontology for diverse \emph{weakness types} a claim may exhibit, informed by a pilot study with domain experts. 

We leverage \dataset to benchmark cutting-edge multimodal and reasoning LLMs (\texttt{GPT-4o, Gemini-2.0-Flash, o3-mini, o1}) on a suite of claim-centric reviewing tasks, including (1) associating reviewer-identified weaknesses with the paper claims they dispute; (2) predicting fine-grained weakness types for claims and refining reviewer-written weaknesses for enhanced specificity; and (3) verifying claims from scratch using grounded reasoning. Through extensive experiments, we find that even frontier models exhibit significant limitations (except for the label prediction sub-task in (2)) as scientific claim verifiers and reviewing assistants. To summarize our contributions, we:
\begin{enumerate}[nolistsep]
\item Introduce \dataset, a dataset of real-world scientific papers, \textit{claim-grounded} reviews, and rich expert annotations;
\item Present a novel suite of tasks for scientific claim verification and \textit{claim-centric} paper review evaluation, enabled by \dataset;
\item Report experimental results on these tasks with multimodal and reasoning LLMs, demonstrating shortcomings of current models for automated, claim-grounded peer review.
\end{enumerate}

\section{Related Work}
\label{sec:background}
\paragraph{Automated Peer Review} Automated peer review is a broad and rapidly growing area of research within AI and NLP, encompassing a wide array of tasks and datasets. We refer the reader to \citet{staudinger-etal-2024-analysis} for a general overview and highlight more narrowly relevant work below.

In focusing on grounding reviewer weaknesses in targeted claims, we follow several prior works that emphasize the \emph{dialectic} nature of peer review, in which authors and reviewers respond directly to one another. \citet{cheng-etal-2020-ape} introduce the RR (Review-Rebuttal or APE) dataset for mining arguments from reviews and rebuttals of ICLR submissions, extracting aligned review-rebuttal argument pairs. The ARIES dataset from \citet{darcy-etal-2023-aries} features reviewer comments from submissions to several computer science conferences, automatically aligned to paper edits made in response. \citet{kumar-etal-2023-reviewers}  study disagreements \emph{among reviewers}, introducing the ContraSciView dataset, which contains pairs of reviews from ICLR and NeurIPS, annotated for reviewer contradictions and disagreements. Lastly, \citet{ruggeri-etal-2023-dataset} present ArgSciChat, a dataset of information-seeking dialogues about a small set of NLP papers, curated by having experts exchange questions and answers about each paper, with answers linked to rationale passages in the text.

\paragraph{Claim Verification} Weaknesses identified by reviewers can be understood as \emph{verifying} the claims that they target, and \emph{claim} (or \emph{fact}) \emph{verification} is its own active research program. Historically, datasets and shared tasks for claim verification, such as FEVER \citep{ thorne-etal-2018-fever, thorne-etal-2018-fact}, SCIVER \citep{wadden-lo-2021-overview}, COVID-Fact \citep{saakyan-etal-2021-covid} and AVeriTeC \citep{schlichtkrull-etal-2024-averitec}, have tended to emphasize prediction of scalar veracity judgments over written explanations \citep[as weaknesses provide;][]{dmonte-etal-2024-claim}, although a number of more recent works have given more attention to the latter \citep[][\emph{i.a.}]{yang-etal-2022-coarse, rani-etal-2023-factify, ma-etal-2024-ex}.

Beyond SCIVER and COVID-Fact, several other claim verification datasets focus on scientific domains. Notable examples include SciFACT \citep{wadden-etal-2020-fact}, which features 1.4k expert-written scientific claims from a variety of fields (e.g.\ microbiology, public health); SciFACT-Open \citep{wadden-etal-2022-scifact}, which builds on SciFACT, with an additional 279 claims from similarly diverse areas; and SciTAB \citep{lu-etal-2023-scitab}, which provides a set of 1.2k claims describing table results extracted from arXiv papers on computer science, requiring compositional reasoning on tables for their verification \citep{sarrouti-etal-2021-evidence-based, wang-etal-2021-semeval, akhtar-etal-2022-pubhealthtab}. We refer the reader to \citet{dmonte-etal-2024-claim} for a general overview of claim verification.

\paragraph{Our Work vs. Prior Work} While \dataset draws raw data from similar sources as other works on automated peer review(\emph{viz.}\ NeurIPS OpenReview submissions), it is unique in focusing on the relationship between reviewer-identified weaknesses and papers' claims.

Within the claim verification literature, our work is distinctive in \emph{drawing evidence for disputed claims from reviews} and in \emph{leveraging complete paper data} (text, images, figures, algorithms, captions)---both from the reviewed paper and from related works---for verification.

Finally, the suite of tasks enabled by \dataset that we explore in \S\ref{sec:experiments} are novel for both domains surveyed above. For automated peer review, these tasks introduce a novel \emph{claim-centric} focus to review generation and evaluation. And for claim verification, these tasks provide a more realistic setting for verification of ``in-the-wild'' claims taken from real papers, leveraging fine-grained weakness types rather than binary veracity labels.

\section{\dataset Construction}
\label{sec:data}

Given a paper and one of its reviews, we aim to collect pairs consisting of (1) a \emph{claim-related weakness} and (2) one or more \emph{target claims}. We define a \emph{{claim-related weakness}} as a contiguous passage from the review that disputes the validity of one or more claims that the paper makes.\footnote{Claim-related weaknesses can be contrasted with those \emph{not} about (a) specific claim(s) made in the paper, such as those highlighting key omissions or issues with the paper taken as a whole. Such weaknesses are not the focus of our work.} For each weakness we also collect a detailed set of labels.

We first describe our data preprocessing pipeline (\S\ref{sec:data::preprocessing}), followed by our annotation tasks (\S\ref{sec:data::tasks}) and the actual annotation process (\S\ref{sec:data::annotation}).

\subsection{Data Sourcing and Preprocessing}
\label{sec:data::preprocessing}
In selecting papers and reviews for \dataset annotation, we sought a corpus that satisfied the following desiderata: (1) \emph{open-access}: the papers and reviews should be publicly available; (2) \emph{domain}: paper topics should align with the expertise of our annotators (primarily NLP); (3) \emph{recency}: the papers should reflect relatively up-to-date research trends in AI and NLP; and (4) \emph{version alignment}: the publicly available versions of the papers should be the \emph{exact} version that the reviews comment on.

After an initial search, we found that rejected OpenReview submissions to NeurIPS 2023 and 2024 met these criteria. 
We exclude \emph{accepted} submissions because only the camera-ready versions of these manuscripts are publicly accessible---\emph{not} the earlier versions that received reviews.

We obtain an initial set of 1,575 publicly available reviews (from 378 rejected papers) from the OpenReview API,\footnote{\url{https://docs.openreview.net/reference/api-v2}} which is then filtered using a two-step process. First, we select reviews that contain at least one of a predefined set of claim-related keywords (see \autoref{app:dataset}). We then further filter this subset to reviews of papers that are broadly related to NLP---our annotators' primary area of expertise---determined by zero-shot prompting GPT-4o.\footnote{\url{https://openai.com/index/hello-gpt-4o/}} This process yielded a final set of 60 reviews and 41 papers for annotation.
We download the PDFs for all 41 papers and parse the full text using PaperMage \citep{lo-etal-2023-papermage}, and further clean the text to mitigate OCR errors. 
We then manually crop all tables, figures, and algorithms as images, along with the captions for each. For LLMs lacking vision capabilities (e.g. \texttt{o3-mini}), we additionally provide dense captions for all images as alternative (textual) inputs.
Finally, we automatically extract claims from the full paper text. Text cleaning, topic classification, (dense) captioning and claim extraction are all done by zero-shot prompting GPT-4o (see \autoref{app:prompts} for prompts).

Finally, a number of the reviews cite related works in connection with the issues they raise. These works thus often provide information critical to assessing the review and the claim(s) it disputes. To ensure that these works are included, we manually read through each review, identifying related works that they cite, and then perform the same preprocessing steps described above on each. This process yielded 56 related work papers.

\begin{figure*}[h]
    \centering
    \includegraphics[width=\linewidth]{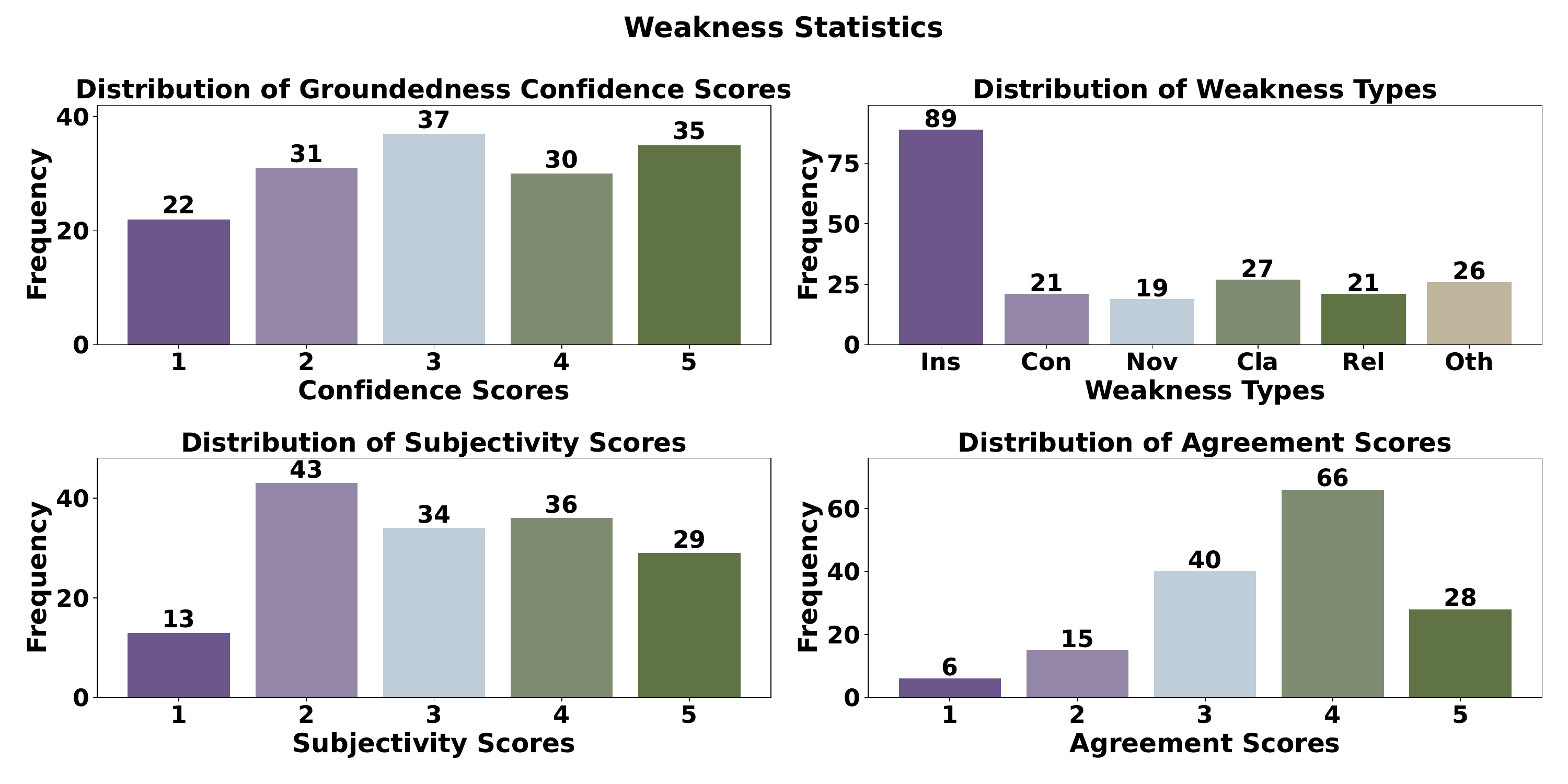}\vspace{-5mm}
    \caption{Distribution of the various weakness labels for \dataset: groundedness confidence scores (top left), weakness types (top right), subjectivity scores (bottom left), and agreement scores (bottom right). See \S\ref{sec:data::annotation}.}
    \label{fig:groundedness-labels}
\end{figure*}

\subsection{Annotation Tasks}
\label{sec:data::tasks}
\dataset annotation consists of three tasks:
\begin{enumerate}[noitemsep]
    \item \textbf{Weakness Identification (WI)}: Identifying \emph{claim-related weaknesses} in reviews.
    \item \textbf{Claim Association (CA)}: Identifying \emph{target claims} in papers disputed by each weakness.
   \item \textbf{Weakness Labeling (WL)}: Providing a set of informative labels for each weakness.
\end{enumerate}
All three tasks take as input the full paper PDF and a single review of that paper. Further task-specific information is provided depending on the task. The WI task was conducted in one annotation interface and the CA and WL tasks were conducted together in another. \autoref{app:annotation} contains screenshots of the interfaces and other annotation information.

\paragraph{Weakness Identification (WI)} involves highlighting contiguous passages in reviews that describe \emph{claim-related weaknesses} (see the first paragraph of \S\ref{sec:data}). We ask annotators to avoid highlighting passages that raise issues that are clearly \emph{not} based on a specific paper claim or result (e.g.\ unclear exposition, missing related work). Annotators then provide a \emph{groundedness} confidence label (1-5) for each weakness, indicating the extent to which they believe the weakness can and should be grounded in an explicit claim in the paper (5), rather than in a broad or speculative claim imputed by the reviewer (1). These labels are not of direct interest, and are collected merely to help annotators in the CA task.

\paragraph{Claim Association (CA)} involves identifying claims in a paper that are \emph{target claims} of the weaknesses annotated in WI. We say that a claim $c$ is a \emph{target claim} of a \emph{claim-related weakness} $w$ iff (1) the truth or accuracy of $c$ is clearly disputed by $w$ and (2) determining this does not require appealing to any other claim(s).\footnote{Condition (2) thus restricts target claims to those \emph{most directly implicated} by the weakness; a weakness in one claim may have implications for others, but we do not count these others as \emph{target claims} for our purposes.} Importantly, not all weaknesses have target claims. This is the rationale for collecting groundedness confidence labels in WI: to help CA annotators triage those that are (not) likely to be groundable in an explicit target claim. In addition to paper and review information, annotators are given the set of claims automatically extracted from the paper by GPT-4o (\S\ref{sec:data::preprocessing}) and are asked to select the target claims from this set. 

\begin{table*}[ht]
    \centering
    \small
    \setlength{\tabcolsep}{5pt}.
    \begin{tabular}{l|ccccccccc}
    \toprule
         &  \textbf{Agr}ement &  \textbf{Sub}jectivity &  \textbf{Con}tradictory &  \textbf{Ins}ufficient &  \textbf{Nov}elty & \textbf{Rel}ated Work & \textbf{Cla}rity &  \textbf{Oth}er \\
         \midrule
         \bf Humans Only & 13.1 & 18.2 & 17.9 & 44.6 & 77.6 & \textbf{52.4} & \phantom{0}0.0 & 22.8 \\
         \midrule
         + \bf \texttt{GPT-4o} & \phantom{0}9.1 & 20.7 & 17.1 & 40.3 & \textbf{78.3} & 46.8 & \phantom{0}2.7 & 17.5 \\
         + \bf \texttt{Gemini-2.0} & 11.6 & 20.2 & 17.2 & 40.4 & 73.9 & 41.4 & \phantom{0}9.3 & \textbf{26.4} \\
         + \bf \texttt{o3-mini} & \textbf{16.5} & 22.6 & 22.3 & 45.6 & 75.2 & 39.4 & \textbf{10.0} & 13.7 \\
         + \bf \texttt{o1} & 14.3 & \textbf{23.3} & \textbf{25.1} & \textbf{48.1} & \textbf{78.3} & 39.3 & \phantom{0}0.2 & 21.1 \\
    \bottomrule
    \end{tabular}
    \caption{Agreement ($\alpha \times$100) on the WL Pilot data between annotators without (top, \S\ref{sec:data::annotation}) and with (bottom, \S\ref{sec:experiments::weakness-editing}) each LLM included as an additional annotator. \textbf{Agr} and \textbf{Sub} are ordinal data (1-5); the rest are binary. At least one of the LLMs improves the average agreement across all the labels, except \textbf{Rel}, and reasoning-enhanced LLMs (\texttt{o1} and \texttt{o3-mini}) bring performance gains across most of the labels.}
    \label{tab:wl-pilot-results}
\end{table*}

Along with the paper, review, and extracted claims, the annotation interface shows the weaknesses identified in WI and the set of extracted candidate claims. Annotators toggle through the claims, tentatively indicating for each whether they think it \emph{may} be a target of one or more weaknesses. Only after seeing all claims do annotators finalize each weakness's target claims by selecting a subset of the ones tentatively identified so far. Additionally, if annotators feel that a weakness clearly targets \emph{some} claim in the paper---but one absent from the candidate set (due to an extraction failure of GPT-4o)---they are instructed to enter it manually.

\paragraph{Weakness Labeling (WL)} asks annotators to label weaknesses given the pairs of claim-related weaknesses and their target claims collected from WI and CA.  
The additional labels include: (1) an ordinal \emph{subjectivity} rating, indicating to what extent the weakness is based on subjective factors (e.g.\ interest in the topic) vs.\ objective facts; (2) an ordinal \emph{agreement} rating, indicating to what extent the annotator finds the weakness well-founded; and (3) one or more \emph{weakness type} labels, characterizing the issue(s) the weakness raises about the claim(s): \emph{insufficient evidence}, \emph{contradictory evidence}, \emph{novelty}, \emph{clarity}, \emph{related work}, or \emph{other}.\footnote{Both the weakness taxonomy and the decision to have a multi-label (vs.\ categorical) scheme were determined by the annotation team through multiple rounds of reading papers and reviews prior to the main \dataset annotation.}

\subsection{Annotation Process}
\label{sec:data::annotation}
All annotators are authors of this work and are either Ph.D. students or full-time researchers in AI/NLP. None received monetary compensation.

\paragraph{WI: Pilot} Pilot annotations for this task were collected on a set of five (paper, review) pairs. Six annotators completed the WI pilot. We calculate pairwise agreement between annotators on weakness span selection by (1) aligning their weakness spans via maximum bipartite matching, using normalized \emph{edit} distance as the span similarity; then (2) computing micro-average pairwise span $\text{F}_1$ using this same similarity (in lieu of exact match) given that alignment, obtaining $\text{F}_{1,edit} = 52.4$.\footnote{We use edit distance rather than exact match for span $\text{F}_1$ given that annotators may exhibit minor differences in how they determine span extents.}

\paragraph{WI: Main} All main annotation examples were singly annotated. Five of the six annotators from the WI pilot performed this annotation and were instructed to annotate up to 20 reviews each. In total, we obtain 168 weaknesses across the 60 reviews. \autoref{fig:groundedness-labels} shows the distributions of groundedness confidence scores, weakness types, subjectivity scores, and agreement scores in \dataset.

\paragraph{WL + CA: Pilot} Five annotators completed a pilot for the WL and CA subtasks (both done in the same interface) using the same set of five (paper, review) pairs as in the WI pilot. The input weaknesses were drawn from the WI pilot annotations of the annotator with the highest $\text{F}_{1,edit}$ agreement. Similar to the above, we report $\text{F}_{1,edit}$ on the identified target claims, obtaining a value of $45.8$. Since annotators are also largely choosing from among a fixed set of candidate claims (rather than unrestricted span selection, as in WI), we further report exact-match $\text{F}_1$, obtaining $\text{F}_{1,exact} = 28.5$. 

We report Krippendorff's $\alpha$ \citep{krippendorff-1970-estimating} for (1) the weakness type labels, (2) weakness subjectivity, and (3) weakness agreement, using the nominal form of the alpha for each label in (1) and the ordinal form for (2) and (3). Results are shown in the first row of \autoref{tab:wl-pilot-results}. For (1), we observe significant variability in agreement across labels---finding medium-to-high agreement for \textbf{Ins}ufficient evidence ($\alpha_\textbf{Ins}=44.6$), \textbf{Rel}ated work ($\alpha_\textbf{Rel}=52.4$), and \textbf{Nov}elty ($\alpha_\textbf{Rel}=77.6$), but lower agreement on other labels. Some pilot annotators disagreed on the border between merely \textbf{Ins}ufficient and outright \textbf{Con}tradictory evidence, for instance, and issues of ``\textbf{Cla}rity'' are sometimes invoked when the primary issue is something else (e.g.\ novelty: ``It is \emph{unclear} whether the proposed method X differs meaningfully from method Y.'') 

For (2) and (3), we find fairly modest agreement ($\alpha_\textbf{Agr}=18.2$, $\alpha_\textbf{Sub}=13.1$), which is unsurprising, as assessments of the validity of a given weakness (\textbf{Agr}) and its subjectivity (\textbf{Sub}) are themselves subject to a large degree of inter-expert \emph{dis}agreement in real-world peer review settings. More generally, we interpret the variability across (1), (2), and (3) as highly characteristic of the variability in judgments observed among actual reviewers.

\paragraph{WL + CA: Main} All of the annotators from the CA subtask pilot participated in the CA main annotation and were again instructed to annotate no more than 20 reviews. 

\paragraph{Statistics}
In total, we obtained 154 target claims across the 60 reviews, where 120 of 168 weaknesses had at least one target claim. Summary statistics for \dataset are shown in \autoref{tab:summary-stats}.

\begin{table}[h]
    \centering
    \small
    \begin{tabular}{lr}
    \toprule
    Data Type & Count \\ 
    \midrule
        Papers & 41 \\
        Reviews & 60 \\
        Related Work Papers & 56 \\
        Target Claims & 154 \\
        Weaknesses & 168 \\
        $\rightarrow$ w/ Target Claims & 120 \\
    \bottomrule
    \end{tabular}
    \caption{Summary statistics for \dataset.}
    \label{tab:summary-stats}
\end{table}

\section{Experiments}
\label{sec:experiments}
\begin{figure*}[ht]
    \centering
    \includegraphics[width=\linewidth]{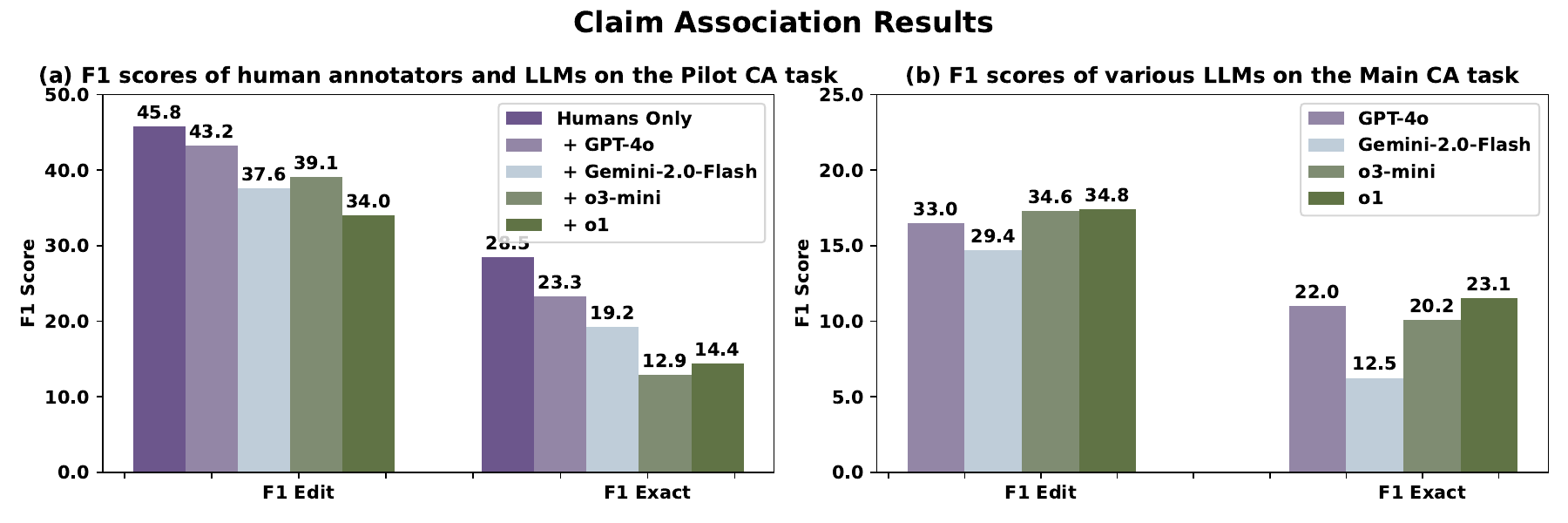}
    \caption{The results of the Claim Association (CA) task (\S\ref{sec:experiments::claim-association}). Left: Avg.\ pairwise $\text{F}_{1,edit}$ and $\text{F}_{1,exact}$ (see \S\ref{sec:data::annotation}) between (1) human annotators (\textbf{Humans Only}) and (2) each model and all humans on the Pilot data. Right: Avg.\ model $\text{F}_1$ w.r.t. single human annotation on the Main data. On the Pilot data, all LLMs show lower performance than the expert (human) average. On the Main data, while \texttt{o1} achieves the highest $F_1$ scores, the low absolute scores of all models indicate that the evaluated LLMs all struggle in grounding weaknesses to claims.}
    \label{fig:ca_results}
\end{figure*}

To support progress on claim-grounded review with LLMs, our experiments benchmark a set of cutting-edge multimodal and reasoning models (\texttt{GPT-4o, Gemini-2.0-Flash, o3-mini, o1}) in the zero-shot setting on three sets of experiments that leverage \dataset and that are motivated by specific use cases (described below): Claim Association (CA), Weakness Labeling and Editing (WLE), and Claim Verification (CV).\footnote{Hyperparameters and prompts in Appendices \ref{app:experiments} and \ref{app:prompts}.}

We use the Pilot and Main annotation sets to evaluate the CA tasks and the label prediction of the WLE task, and use Main data only for other tasks. We summarize our findings as follows (\S\ref{sec:experiments::claim-association}-\S\ref{sec:experiments::claim-verification} contain detailed results and analysis):
\begin{itemize}
    \item For the CA task (\S\ref{sec:experiments::claim-association}), all LLMs exhibit poorer agreement with human-only average on the Pilot set (\autoref{fig:ca_results} (a)) and with (single) human annotation on the Main set (\autoref{fig:ca_results} (b)), suggesting that they fall short in grounding review weaknesses to paper claims.
    \item For the WLE task (\S\ref{sec:experiments::weakness-editing}), LLM-predicted weakness labels for target claims demonstrate similar or better agreement than human-only average on the Pilot set (\autoref{tab:wl-pilot-results}) and moderate to high level alignment with single human annotation on the Main set (\autoref{tab:we-label-results}). The results showcase their usefulness in aiding humans in supplying fine-grained weakness information about problematic claims when given reviewer-written weaknesses as a reference. However, they still struggle to \emph{edit} these weaknesses for enhanced specificity and groundedness (\autoref{tab:we-span-results}).
    \item For the CV task (\S\ref{sec:experiments::claim-verification}), all LLMs perform poorly in generating claim-grounded weaknesses \emph{de novo} (i.e.\ without the human reference), indicating that verifying scientific claims with paper-grounded reasoning remains challenging even for recent LLMs.
\end{itemize}

\begin{table*}
    \vspace{5mm}
    \centering
    \small
    \setlength{\tabcolsep}{5pt}
    \begin{tabular}{l|cccccccc}
    \toprule
         \, & \textbf{Agr}ement &  \textbf{Sub}jectivity &  \textbf{Con}trast &  \textbf{Ins}ufficient &  \textbf{Nov}elty & \textbf{Rel}ated Work & \textbf{Cla}rity &  \textbf{Oth}er \\
         \midrule
          & \multicolumn{8}{c}{\bf Weakness Labeling (with review weaknesses as references)} \\
        \midrule
        \bf \texttt{GPT-4o} & 25.6 & 19.5 & 24.6 & 52.5 & 74.4 & 28.0 & 31.9 & \phantom{0}0.4
        \\
        \bf \texttt{Gemini-2.0} & 14.4 & 23.9 & 41.4 & 62.9 & 73.4 & \textbf{46.6} & 45.5 & \textbf{16.6} 
        \\
        \bf \texttt{o3-mini} & 27.2 & 26.2 & 46.5 & 67.2 & 88.8 & 39.7 & 46.7 & 14.6 
        \\
        \bf \texttt{o1} & \textbf{27.3} & \textbf{30.4} & \textbf{51.0} & \textbf{70.6} & \textbf{94.9} & 43.9 & \textbf{49.8} & \phantom{0}9.6 
        \\
        \midrule
          & \multicolumn{8}{c}{\bf Claim Verification (without review weaknesses as references)} \\
        \midrule
        \bf \texttt{GPT-4o} & - & - & \,-5.4 & \phantom{0}4.6 & \textbf{27.9} & \textbf{18.5} & \,-5.6 & \phantom{0}0.4
        \\
        \bf \texttt{Gemini} & - & - & \textbf{\phantom{0}6.8} & \textbf{23.4} & 24.9 & 15.4 & \textbf{\phantom{0}2.4} & \textbf{\phantom{0}1.7}
        \\
        \bf \texttt{o3-mini} & - & - & \,-0.1 & \,-6.6 & 27.8 & \phantom{0}8.2 & \,-11.6 & \,-7.8 
        \\
        \bf \texttt{o1} & - & - & \phantom{0}6.6 & \phantom{0}8.7 & 12.8 & \,-5.4 & \,-4.3 & \,-2.6 
        \\
    \bottomrule
    \end{tabular}
    \caption{Agreement ($\alpha \times$100) between LLMs and gold \dataset agreement scores (\textbf{Agr}), subjectivity scores (\textbf{Sub}), and weakness type labels on the main set for \textbf{Top}: weakness labeling task (\S\ref{sec:experiments::weakness-editing}), and \textbf{Bottom}: claim verification task (\S\ref{sec:experiments::claim-verification}). On the weakness labeling task, \texttt{Gemini-2.0-Flash} achieves the best agreement with human on \textbf{Rel} and \textbf{Oth}, and \texttt{o1} performs the best on all the remaining labels. The results demonstrate the capability of LLMs to predict fine-grained weakness categories for claims given reviews as references. On the claim verification task, all LLMs perform significantly worse on all the labels, compared with when they are provided with human-written weaknesses in the WLE task (the corresponding cells in the top section of the table), suggesting the evaluated LLMs struggle to verify and identify the weakness of scientific claims from scratch.}
    \label{tab:we-label-results}
\end{table*}

\begin{table*}[t]
    \centering
    \begin{tabular}{p{0.47\textwidth}p{0.47\textwidth}}
    \bf Original (Human) & \bf Revised (GPT-4o) \\
    \midrule
        \small \textcolor{original}{In Table 2, I would like to see the performance of other correction methods (e.g., GPT3.5/4/4o) for a more comprehensive comparison.} & \small \textcolor{revised}{In Table 2, the performance of other correction methods (e.g., GPT3.5/4/4o) should be included to provide a more comprehensive comparison and substantiate the claims of scalability and generalizability of the proposed solution.} \\
        \midrule
        \small \textcolor{original}{Line 78: There should be some statistical significance by which the models perform poorly and then you could consider it underperformance and not just lower than actual. Models being non-deterministic, there is always a chance of slight changes in scores.} & \small \textcolor{revised}{The paper should include statistical significance testing to determine if the models' underperformance is truly indicative of sandbagging rather than random variation. Given the non-deterministic nature of AI models, this would provide stronger evidence for the claims made.} \\
        \midrule
        \small \textcolor{original}{What evidence (empirical results) can support the claim that 'the combination of both improvements --- text splitting and iterated calls, has proven itself to perform the best' (line 146)} & \small \textcolor{revised}{The paper lacks empirical evidence to support the claim that the combination of text splitting and iterated calls performs best. It would benefit from experiments or data demonstrating this improvement, such as comparative analysis with other methods or detailed performance metrics.} \\
    \bottomrule
    \end{tabular}
    \caption{Examples of (original, revised) weakness pairs from the weakness editing task (\S\ref{sec:experiments::weakness-editing}) where GPT-4o (\textbf{Revised}) fails to improve upon the specificity of the human-written (\textbf{Original}) weakness---a common occurrence in our human evaluation (See \S\ref{sec:experiments::weakness-editing}).}
    \label{tab:we-span-examples}
\end{table*}

\subsection{Claim Association (CA)}
\label{sec:experiments::claim-association}
The CA task is motivated by a scenario in which a reviewer has identified a weakness with a paper and would like an LLM to help collect in-text citations to ground it. We provide models with a single claim-related weakness, the paper contents, and the same set of candidate claims and instructions as were given to annotators (see \S\ref{sec:data::tasks}), asking the model to identify up to three target claims for the provided weakness. Models were also permitted to supply a custom claim, as in \S\ref{sec:data::tasks}.

\paragraph{Results} As in \S\ref{sec:data::annotation}, we report $\text{F}_{1,edit}$ and $\text{F}_{1,exact}$ to assess LLM agreement with human annotators. \autoref{fig:ca_results} (left) shows average pairwise $\text{F}_1$ scores (1) among humans only and (2) between each LLM and all humans on the \dataset pilot data. LLMs exhibit consistently lower average scores than the human average.

\autoref{fig:ca_results} (right) shows average $\text{F}_1$'s between each model and the (singly annotated) main \dataset data. Here, we find \texttt{o1} achieves the strongest results, though \texttt{o3-mini} and \texttt{GPT-4o} are comparable. In absolute terms, scores are low across the board, suggesting that models struggle to identify appropriate target claims. 

\subsection{Weakness Labeling and  Editing (WLE)}
\label{sec:experiments::weakness-editing}
This task is motivated by the needs of \emph{meta-reviewers} who must synthesize primary reviews. We envision that an LLM may be used to \emph{enrich} primary reviews by providing weakness labels and by enhancing their specificity, helping the meta-reviewer more efficiently write their own review.

We provide an LLM with the full contents of the reviewed paper (text, tables, figures, images, and captions), a reviewer-written weakness, its target claims, and the full contents of related work(s) mentioned either by the target claim(s) or the weakness. We ask the model to provide weakness types and agreement and subjectivity scores and---if the model deems necessary---an \emph{edited} weakness that enhances the specificity and groundedness of the original. We evaluate only on \emph{claim-related} weaknesses (i.e.\ ones with $\geq 1$ target claim).

\paragraph{Results: Labels}
We first consider the model-predicted annotations for weakness type (\textbf{Con}tradictory evidence, \textbf{Ins}ufficient evidence, \textbf{Nov}elty, \textbf{Rel}ated Work, \textbf{Cla}rity, \textbf{Oth}er) and for \textbf{Agr}eement, and \textbf{Sub}jectivity. As with CA, we evaluate each LLM as an additional annotator for the Pilot data, and with the single human annotation for the Main data. Tables \ref{tab:wl-pilot-results} and \ref{tab:we-label-results} report $\alpha$ on the Pilot and Main data, respectively.

On the Pilot data (\autoref{tab:wl-pilot-results}), at least one LLM boosts overall agreement ($\alpha$) for each label (except \textbf{Rel}) relative to the human-only annotator set, and some LLMs---notably \texttt{o1} and \texttt{o3-mini}---raise $\alpha$ across numerous labels.

On the Main data (\autoref{tab:we-label-results}, top section), we observe strong agreement for \textbf{Nov} and \textbf{Ins}, and a moderate-to-strong level of agreement for \textbf{Rel}, consistent with the Pilot subset. This is intuitive, as weaknesses of these kinds are often readily identifiable from common lexical cues (e.g. \emph{novel(ty)}, \emph{convincing}) and explicit citations (for \textbf{Rel}). LLMs exhibit similar or greater agreement for other labels (\textbf{Con}, \textbf{Rel}, \textbf{Cla}, \textbf{Agr}, and \textbf{Sub}) relative to the Pilot data, and similar or lower for \textbf{Oth}. \texttt{Gemini-2.0-Flash} attains the best overall performance on \textbf{Rel} and \textbf{Oth}, and \texttt{o1} performs best on all the remaining labels.

Overall, these results suggest that recent LLMs---especially reasoning models like \texttt{o1}---may be able to act as useful aids to metareviewers in offering fine-grained information about reviewer-identified weaknesses to be synthesized in the metareview.

\paragraph{Results: Edited Weaknesses}
Next, we compare the \emph{texts} of the revised weaknesses with those of the original. We evaluate models' ability to enhance the specificity and groundedness of reviewer-written weaknesses by conducting a human evaluation of 20 randomly sampled examples from the Main data. For each example, two human judges are provided with the original weakness, the model-revised weakness, and the target claims, with the provenance of the two weaknesses (model vs.\ human) hidden and presentation order randomized. The judges indicate which weakness they believe to be more specific and grounded, with ties permitted. Judges also indicate whether the core meaning of the two weaknesses is the same or different.

\autoref{tab:we-span-results} reports the average proportion of cases in which a judge preferred the (original) \textbf{Human}-written weakness, the \textbf{Model}-revised one, or neither (\textbf{Tie}). \textbf{Error} represents examples where the core meaning of the model-revised weakness was deemed different from the original. We find that models generally fail to improve upon the original weaknesses---with ties consistently the plurality (if not majority) judgment across models---though \texttt{o3-mini} manages to enhance weaknesses in a meaningful fraction of cases (30.0\%). On further inspection, we find that models tend to make revisions that render the tone of the review more polite (e.g.\ by moving from first- to third-person), or that verbalize a suggestion already strongly implied in the original review, without actually providing more concrete feedback (\autoref{tab:we-span-examples}, \emph{top}). Sometimes, models even strip out helpful textual anchors, such as line numbers and quotation marks (\autoref{tab:we-span-examples}, \emph{middle}, \emph{bottom}), making it more difficult to locate the disputed claim, and thus making the revised weakness \emph{less} grounded.

\begin{table}
    \centering
    \small
    \begin{tabular}{l|cccc}
    \toprule
        & \bf Human\% & \bf Model\% & \bf Tie\% & \bf Error\%  \\
    \midrule
        \bf \texttt{GPT-4o} & 20.0 & 12.5 & 72.5 & \phantom{0}5.0 \\
        \bf \texttt{Gemini} & \phantom{0} 2.5 & 15.0 & 77.5 & \phantom{0}5.0\\
        \bf \texttt{o3-mini} & \phantom{0}7.5 & 30.0 & 45.0 & 17.5 \\
        \bf \texttt{o1} & 30.0 & 22.5 & 45.0 & \phantom{0}2.5 \\
    \bottomrule
    \end{tabular}
    \caption{Avg.\ \% cases ($N = 20/\text{model}$) in which two human judges deemed the \textbf{Human}-written weakness most \emph{specific} and \emph{grounded}, the \textbf{Model}-revised one, or neither (\textbf{Tie}) in our weakness editing task (\S\ref{sec:experiments::weakness-editing}). \textbf{Error} captures cases where the model-revised weaknesses were judged to have changed the original meaning. While \texttt{o3-mini} can enhance human-written weaknesses for a moderate portion of the cases, other LLMs generally fail to provide improvements over the original human-written weakness.}
    \label{tab:we-span-results}
\end{table}

\subsection{Claim Verification (CV)}
\label{sec:experiments::claim-verification}
This task requires a model to verify claims made in the paper \emph{from scratch} by providing their own claim-grounded weakness---not relying on a reviewer-identified weakness (see \S\ref{sec:experiments::weakness-editing}). This is precisely what human reviewers are required to do in assessing a paper's claims. Our evaluation for this task compares model-generated weaknesses for a \emph{single} target claim---the \emph{focal} claim---to reviewer-identified weaknesses of the same claim.

We obtain model-generated weaknesses for a given focal claim by providing the claim as input in a prompt, along with the details of the paper and related work(s) as in previous experiments. The prompt asks the model to (1) extract pieces of evidence from the paper(s) needed to assess the focal claim; (2) describe a weakness of that claim; and (3) perform weakness labeling on the result.

Since each reviewer-identified weaknesses $w$ in \dataset may be associated with up to three target claims ($t_1$, $t_2$, $t_3$), we use a model  to summarize the part(s) of $w$ related to each of its target claims in turn---yielding up to three distilled weaknesses ($w'_1, w'_2, w'_3$), each related to a single focal claim, to be used as the human references for the corresponding model-generated weaknesses.

To perform this distillation, we provide GPT-4o with a prompt containing the original weakness ($w$), a focal claim ($t_i$), and details of the paper and related work(s). The prompt asks the model to (1) extract pieces of evidence from the paper(s) needed to assess the focal claim and (2) provide the distilled weakness based on the focal claim and this evidence. We then take (2) as the reviewer's weakness for the focal claim.

Finally, given the reference and model-generated weaknesses, we (1) compare the model-generated weakness labels with those of the reference weakness and (2) use an LLM-based evaluation to determine whether these weaknesses describe exactly the \emph{same} issue, merely \emph{similar} issues, or entirely \emph{different} issues with the focal claim. We provide the focal claim and the two weaknesses as input to the evaluation prompt, along with the evidence extracted for each weakness in the previous steps.

\paragraph{Results: Weakness Text} We use GPT-4o as the LLM judge. As shown in \autoref{tab:cv-text-results}, we find that LLM-generated weaknesses across all evaluated models for the focal claim overwhelmingly tend to be judged \emph{different} from those identified by the reviewers ($> 80\%$ of cases). A smaller portion of these reviews are deemed \emph{similar} to those of the reviewers ($\sim 10\%$), and an even smaller fraction are considered the same ($<10\%$). While \emph{different} here does not necessarily mean \emph{wrong}, manual inspection reveals that model-written weaknesses tend to be overly generic in their diagnoses (e.g.\ \emph{``there is a lack of precise evidence linking GSNR to controlling the generalization gap as claimed''}) and sometimes make more basic errors, such as denying that the paper comments on the claim at all. These results indicate that writing specific, claim-grounded weaknesses from scratch remains a challenge even for contemporary models.

\begin{table}
    \centering
    \small
    \adjustbox{max width=\columnwidth}{
    \begin{tabular}{l|lll}
    \toprule
        & \bf Same\% & \bf Sim\% & \bf Diff\% \\
    \midrule
        \bf \texttt{GPT-4o} & \phantom{0}6.8 & 11.9 & 81.0 \\
        \bf \texttt{Gemini-2.0} & \phantom{0}4.8 & \phantom{0}9.3 & 85.9 \\
        \bf \texttt{o3-mini} & \phantom{0}9.3 & 10.2 & 80.5 \\
        \bf \texttt{o1} & \phantom{0}8.7 & \phantom{0}8.0 & 83.0 \\
    \bottomrule
    \end{tabular}
    }
    \caption{Model-generated weaknesses judged by \texttt{GPT-4o} to be the same as (\textbf{Same}), similar to (\textbf{Sim}), or different from (\textbf{Diff}), the distilled reviewer-identified ones on our claim verification (CV) task (\S\ref{sec:experiments::claim-verification}). The LLM-generated weakness are overwhelmingly judged to be \emph{different} from reviewer-identified weakness.}
    \label{tab:cv-text-results}
\end{table}

\paragraph{Results: Labels} Consistent with the comparison of weakness texts above (\autoref{tab:cv-text-results}), we find in the bottom section of \autoref{tab:we-label-results} that the weakness \emph{types} assigned to the reviewer-identified and model-generated weaknesses diverge as well---showing uniformly lower agreement across types than was observed in the weakness editing task (\autoref{tab:we-label-results}, top). This  suggests the model-generated weaknesses tend to differ \emph{in kind} from the reviewer-written ones, and reveal the inability of the leading LLMs to verify and identify the weakness of scientific claims from scratch without human-written weaknesses as references.

\section{Conclusion}
\label{sec:conclusion}
This work has introduced \dataset---a benchmark of reviewer-identified weaknesses in NeurIPS 2023 and 2024 submissions, richly annotated with descriptive labels by experts and grounded in the \emph{claims} that they dispute in the reviewed papers. Further, we benchmark various LLMs on three novel tasks enabled by \dataset---Weakness Labeling and Editing (WLE), Claim Association (CA), and Claim Verification (CV)---all aimed at assisting reviewers during the peer review process. Across these tasks, we find even cutting-edge LLMs struggle to provide specific, grounded reviews and to identify and verify the specific claims targeted by those reviews. We release \dataset to support further research on claim-grounded automated peer review.

\section*{Limitations}
\label{sec:limitations}
\dataset focuses on reviewer-identified weaknesses that are  \emph{claim-related}, meaning that they take issue with a particular claim or claim(s) a paper makes. While we believe this kind of weakness is among the most valuable in the peer review process, other kinds can be valuable as well. For example, weaknesses that identify important experiments or related work that were \emph{omitted} are also valuable. Weaknesses of this sort are arguably even harder to identify than our claim-related weaknesses, and empowering models to highlight such cases is an interesting direction for future work.

Second, \dataset is somewhat limited in its size due to (1) the limited sources that satisfy our four desiderata (see \S\ref{sec:data}), and (2) the labor involved in collecting these expert annotations. As such, \dataset is intended purely as an \emph{evaluation} benchmark for LLMs and LLM-based models for peer review, and is likely not large enough for meaningful supervised fine-tuning.

\section*{Ethics}
\label{sec:ethics}
We do not believe this work raises any significant ethical concerns. In collecting \dataset, we have complied with OpenReview licensing and terms of use. Further, since both the papers and the reviews in \dataset are anonymized, there is little concern about leakage of personally identifiable information (PII).

\section*{Acknowledgment}
This material is based upon work supported by Defense Advance Research Projects Agency (DARPA) under Contract No. HR001125C0304 and ONR grant (N0001424-1-2089). 
Any opinions, findings and conclusions or recommendations expressed in this material are those of the author(s) and do not necessarily reflect the views of DARPA.
We sincerely thank Jack Zhang, Tianjian Li and Hannah Gonzalez for their constructive feedback on an earlier version of this document.

%

\begin{thebibliography}{25}
\providecommand{\natexlab}[1]{#1}

\bibitem[{Akhtar et~al.(2022)Akhtar, Cocarascu, and Simperl}]{akhtar-etal-2022-pubhealthtab}
Mubashara Akhtar, Oana Cocarascu, and Elena Simperl. 2022.
\newblock \href {https://doi.org/10.18653/v1/2022.findings-naacl.1} {{P}ub{H}ealth{T}ab: {A} public health table-based dataset for evidence-based fact checking}.
\newblock In \emph{Findings of the Association for Computational Linguistics: NAACL 2022}, pages 1--16, Seattle, United States. Association for Computational Linguistics.

\bibitem[{Cheng et~al.(2020)Cheng, Bing, Yu, Lu, and Si}]{cheng-etal-2020-ape}
Liying Cheng, Lidong Bing, Qian Yu, Wei Lu, and Luo Si. 2020.
\newblock \href {https://doi.org/10.18653/v1/2020.emnlp-main.569} {{APE}: Argument pair extraction from peer review and rebuttal via multi-task learning}.
\newblock In \emph{Proceedings of the 2020 Conference on Empirical Methods in Natural Language Processing (EMNLP)}, pages 7000--7011, Online. Association for Computational Linguistics.

\bibitem[{D'Arcy et~al.(2023)D'Arcy, Ross, Bransom, Kuehl, Bragg, Hope, and Downey}]{darcy-etal-2023-aries}
Mike D'Arcy, Alexis Ross, Erin Bransom, Bailey Kuehl, Jonathan Bragg, Tom Hope, and Doug Downey. 2023.
\newblock Aries: A corpus of scientific paper edits made in response to peer reviews.
\newblock \emph{arXiv preprint arXiv:2306.12587}.

\bibitem[{Dmonte et~al.(2024)Dmonte, Oruche, Zampieri, Calyam, and Augenstein}]{dmonte-etal-2024-claim}
Alphaeus Dmonte, Roland Oruche, Marcos Zampieri, Prasad Calyam, and Isabelle Augenstein. 2024.
\newblock Claim verification in the age of large language models: A survey.
\newblock \emph{arXiv preprint arXiv:2408.14317}.

\bibitem[{Estornell et~al.(2020)Estornell, Das, and Vorobeychik}]{estornell2020deception}
Andrew Estornell, Sanmay Das, and Yevgeniy Vorobeychik. 2020.
\newblock Deception through half-truths.
\newblock In \emph{Proceedings of the AAAI Conference on Artificial Intelligence}, volume~34, pages 10110--10117.

\bibitem[{Jiang et~al.(2023)Jiang, Liu, Ding, Guo, and Lin}]{jiang-etal-2023-accelerating}
Guo-qing Jiang, Jinlong Liu, Zixiang Ding, Lin Guo, and Wei Lin. 2023.
\newblock Accelerating large batch training via gradient signal to noise ratio (gsnr).
\newblock \emph{arXiv preprint arXiv:2309.13681}.

\bibitem[{Krippendorff(1970)}]{krippendorff-1970-estimating}
Klaus Krippendorff. 1970.
\newblock Estimating the reliability, systematic error and random error of interval data.
\newblock \emph{Educational and psychological measurement}, 30(1):61--70.

\bibitem[{Kumar et~al.(2023)Kumar, Ghosal, and Ekbal}]{kumar-etal-2023-reviewers}
Sandeep Kumar, Tirthankar Ghosal, and Asif Ekbal. 2023.
\newblock \href {https://doi.org/10.18653/v1/2023.emnlp-main.1038} {When reviewers lock horns: Finding disagreements in scientific peer reviews}.
\newblock In \emph{Proceedings of the 2023 Conference on Empirical Methods in Natural Language Processing}, pages 16693--16704, Singapore. Association for Computational Linguistics.

\bibitem[{Lo et~al.(2023)Lo, Shen, Newman, Chang, Authur, Bransom, Candra, Chandrasekhar, Huff, Kuehl, Singh, Wilhelm, Zamarron, Hearst, Weld, Downey, and Soldaini}]{lo-etal-2023-papermage}
Kyle Lo, Zejiang Shen, Benjamin Newman, Joseph Chang, Russell Authur, Erin Bransom, Stefan Candra, Yoganand Chandrasekhar, Regan Huff, Bailey Kuehl, Amanpreet Singh, Chris Wilhelm, Angele Zamarron, Marti~A. Hearst, Daniel Weld, Doug Downey, and Luca Soldaini. 2023.
\newblock \href {https://doi.org/10.18653/v1/2023.emnlp-demo.45} {{P}aper{M}age: A unified toolkit for processing, representing, and manipulating visually-rich scientific documents}.
\newblock In \emph{Proceedings of the 2023 Conference on Empirical Methods in Natural Language Processing: System Demonstrations}, pages 495--507, Singapore. Association for Computational Linguistics.

\bibitem[{Lu et~al.(2023)Lu, Pan, Liu, Nakov, and Kan}]{lu-etal-2023-scitab}
Xinyuan Lu, Liangming Pan, Qian Liu, Preslav Nakov, and Min-Yen Kan. 2023.
\newblock \href {https://doi.org/10.18653/v1/2023.emnlp-main.483} {{SCITAB}: A challenging benchmark for compositional reasoning and claim verification on scientific tables}.
\newblock In \emph{Proceedings of the 2023 Conference on Empirical Methods in Natural Language Processing}, pages 7787--7813, Singapore. Association for Computational Linguistics.

\bibitem[{Ma et~al.(2024)Ma, Xu, Wei, Chen, Wang, Liu, Wu, and Wang}]{ma-etal-2024-ex}
Huanhuan Ma, Weizhi Xu, Yifan Wei, Liuji Chen, Liang Wang, Qiang Liu, Shu Wu, and Liang Wang. 2024.
\newblock \href {https://doi.org/10.18653/v1/2024.findings-acl.556} {{EX}-{FEVER}: A dataset for multi-hop explainable fact verification}.
\newblock In \emph{Findings of the Association for Computational Linguistics: ACL 2024}, pages 9340--9353, Bangkok, Thailand. Association for Computational Linguistics.

\bibitem[{Rani et~al.(2023)Rani, Tonmoy, Dalal, Gautam, Chakraborty, Chadha, Sheth, and Das}]{rani-etal-2023-factify}
Anku Rani, S.M Towhidul~Islam Tonmoy, Dwip Dalal, Shreya Gautam, Megha Chakraborty, Aman Chadha, Amit Sheth, and Amitava Das. 2023.
\newblock \href {https://doi.org/10.18653/v1/2023.acl-long.581} {{FACTIFY}-5{WQA}: 5{W} aspect-based fact verification through question answering}.
\newblock In \emph{Proceedings of the 61st Annual Meeting of the Association for Computational Linguistics (Volume 1: Long Papers)}, pages 10421--10440, Toronto, Canada. Association for Computational Linguistics.

\bibitem[{Ruggeri et~al.(2023)Ruggeri, Mesgar, and Gurevych}]{ruggeri-etal-2023-dataset}
Federico Ruggeri, Mohsen Mesgar, and Iryna Gurevych. 2023.
\newblock \href {https://doi.org/10.18653/v1/2023.acl-long.425} {A dataset of argumentative dialogues on scientific papers}.
\newblock In \emph{Proceedings of the 61st Annual Meeting of the Association for Computational Linguistics (Volume 1: Long Papers)}, pages 7684--7699, Toronto, Canada. Association for Computational Linguistics.

\bibitem[{Saakyan et~al.(2021)Saakyan, Chakrabarty, and Muresan}]{saakyan-etal-2021-covid}
Arkadiy Saakyan, Tuhin Chakrabarty, and Smaranda Muresan. 2021.
\newblock \href {https://doi.org/10.18653/v1/2021.acl-long.165} {{COVID}-fact: Fact extraction and verification of real-world claims on {COVID}-19 pandemic}.
\newblock In \emph{Proceedings of the 59th Annual Meeting of the Association for Computational Linguistics and the 11th International Joint Conference on Natural Language Processing (Volume 1: Long Papers)}, pages 2116--2129, Online. Association for Computational Linguistics.

\bibitem[{Sarrouti et~al.(2021)Sarrouti, Ben~Abacha, Mrabet, and Demner-Fushman}]{sarrouti-etal-2021-evidence-based}
Mourad Sarrouti, Asma Ben~Abacha, Yassine Mrabet, and Dina Demner-Fushman. 2021.
\newblock \href {https://doi.org/10.18653/v1/2021.findings-emnlp.297} {Evidence-based fact-checking of health-related claims}.
\newblock In \emph{Findings of the Association for Computational Linguistics: EMNLP 2021}, pages 3499--3512, Punta Cana, Dominican Republic. Association for Computational Linguistics.

\bibitem[{Schlichtkrull et~al.(2024)Schlichtkrull, Guo, and Vlachos}]{schlichtkrull-etal-2024-averitec}
Michael Schlichtkrull, Zhijiang Guo, and Andreas Vlachos. 2024.
\newblock Averitec: A dataset for real-world claim verification with evidence from the web.
\newblock \emph{Advances in Neural Information Processing Systems}, 36.

\bibitem[{Staudinger et~al.(2024)Staudinger, Kusa, Piroi, and Hanbury}]{staudinger-etal-2024-analysis}
Moritz Staudinger, Wojciech Kusa, Florina Piroi, and Allan Hanbury. 2024.
\newblock \href {https://aclanthology.org/2024.sdp-1.24/} {An analysis of tasks and datasets in peer reviewing}.
\newblock In \emph{Proceedings of the Fourth Workshop on Scholarly Document Processing (SDP 2024)}, pages 257--268, Bangkok, Thailand. Association for Computational Linguistics.

\bibitem[{Thorne et~al.(2018{\natexlab{a}})Thorne, Vlachos, Christodoulopoulos, and Mittal}]{thorne-etal-2018-fever}
James Thorne, Andreas Vlachos, Christos Christodoulopoulos, and Arpit Mittal. 2018{\natexlab{a}}.
\newblock \href {https://doi.org/10.18653/v1/N18-1074} {{FEVER}: a large-scale dataset for fact extraction and {VER}ification}.
\newblock In \emph{Proceedings of the 2018 Conference of the North {A}merican Chapter of the Association for Computational Linguistics: Human Language Technologies, Volume 1 (Long Papers)}, pages 809--819, New Orleans, Louisiana. Association for Computational Linguistics.

\bibitem[{Thorne et~al.(2018{\natexlab{b}})Thorne, Vlachos, Cocarascu, Christodoulopoulos, and Mittal}]{thorne-etal-2018-fact}
James Thorne, Andreas Vlachos, Oana Cocarascu, Christos Christodoulopoulos, and Arpit Mittal. 2018{\natexlab{b}}.
\newblock \href {https://doi.org/10.18653/v1/W18-5501} {The fact extraction and {VER}ification ({FEVER}) shared task}.
\newblock In \emph{Proceedings of the First Workshop on Fact Extraction and {VER}ification ({FEVER})}, pages 1--9, Brussels, Belgium. Association for Computational Linguistics.

\bibitem[{Venkat et~al.(2022)Venkat, Richa, Rao, and Das}]{venkat2022liarx}
Sharanya Venkat, Richa, Gaurang Rao, and Bhaskarjyoti Das. 2022.
\newblock Liarx: A partial fact fake news data set with label distribution approach for fake news detection.
\newblock In \emph{Innovations in Computational Intelligence and Computer Vision: Proceedings of ICICV 2021}, pages 221--229. Springer.

\bibitem[{Wadden et~al.(2020)Wadden, Lin, Lo, Wang, van Zuylen, Cohan, and Hajishirzi}]{wadden-etal-2020-fact}
David Wadden, Shanchuan Lin, Kyle Lo, Lucy~Lu Wang, Madeleine van Zuylen, Arman Cohan, and Hannaneh Hajishirzi. 2020.
\newblock \href {https://doi.org/10.18653/v1/2020.emnlp-main.609} {Fact or fiction: Verifying scientific claims}.
\newblock In \emph{Proceedings of the 2020 Conference on Empirical Methods in Natural Language Processing (EMNLP)}, pages 7534--7550, Online. Association for Computational Linguistics.

\bibitem[{Wadden and Lo(2021)}]{wadden-lo-2021-overview}
David Wadden and Kyle Lo. 2021.
\newblock \href {https://aclanthology.org/2021.sdp-1.16/} {Overview and insights from the {SCIVER} shared task on scientific claim verification}.
\newblock In \emph{Proceedings of the Second Workshop on Scholarly Document Processing}, pages 124--129, Online. Association for Computational Linguistics.

\bibitem[{Wadden et~al.(2022)Wadden, Lo, Kuehl, Cohan, Beltagy, Wang, and Hajishirzi}]{wadden-etal-2022-scifact}
David Wadden, Kyle Lo, Bailey Kuehl, Arman Cohan, Iz~Beltagy, Lucy~Lu Wang, and Hannaneh Hajishirzi. 2022.
\newblock \href {https://doi.org/10.18653/v1/2022.findings-emnlp.347} {{S}ci{F}act-open: Towards open-domain scientific claim verification}.
\newblock In \emph{Findings of the Association for Computational Linguistics: EMNLP 2022}, pages 4719--4734, Abu Dhabi, United Arab Emirates. Association for Computational Linguistics.

\bibitem[{Wang et~al.(2021)Wang, Mahajan, Danilevsky, and Rosenthal}]{wang-etal-2021-semeval}
Nancy X.~R. Wang, Diwakar Mahajan, Marina Danilevsky, and Sara Rosenthal. 2021.
\newblock \href {https://doi.org/10.18653/v1/2021.semeval-1.39} {{S}em{E}val-2021 task 9: Fact verification and evidence finding for tabular data in scientific documents ({SEM}-{TAB}-{FACTS})}.
\newblock In \emph{Proceedings of the 15th International Workshop on Semantic Evaluation (SemEval-2021)}, pages 317--326, Online. Association for Computational Linguistics.

\bibitem[{Yang et~al.(2022)Yang, Ma, Chen, Lin, Luo, and Chang}]{yang-etal-2022-coarse}
Zhiwei Yang, Jing Ma, Hechang Chen, Hongzhan Lin, Ziyang Luo, and Yi~Chang. 2022.
\newblock \href {https://aclanthology.org/2022.coling-1.230/} {A coarse-to-fine cascaded evidence-distillation neural network for explainable fake news detection}.
\newblock In \emph{Proceedings of the 29th International Conference on Computational Linguistics}, pages 2608--2621, Gyeongju, Republic of Korea. International Committee on Computational Linguistics.

\end{thebibliography}

\clearpage

\appendix

\section{Dataset Details}
\label{app:dataset}
\subsection{Licensing and Terms of Use}

The papers and reviews included in \dataset are all obtained from OpenReview and our use of them is consistent with the OpenReview terms of use: \url{https://openreview.net/legal/terms}. Upon paper acceptance, we will release \dataset under a [CC-BY 4.0] license, which is also consistent with these terms.



\subsection {Data Preprocessing}
We use \texttt{GPT-4o-2024-08-06} with zero-shot prompting and \texttt{temperature=1.0} for full-text extraction, text cleaning, caption extraction and topic classification. See \autoref{app:prompts} for the respective prompts.

We filter reviews to contain at least one claim-related keywords from the list: (see \textbf{Claim-related Keywords} on the next page.)
\begin{figure*}
\begin{tcolorbox}[colback={lightgray},title={\textbf{Claim-related Keywords}},colbacktitle=white,coltitle=black]

["overclaim", "over-claim", "over claim", "claim", "claims", "claiming", "claimed", "supported", "fully support", "fully supporting", "fully supportive",
"supported", "support", "supporting", "substantiate", "substantiating", "substantiated", "convincing", "convince", "convincingly", "convinces", "supportive", "unsubstantiated", "unsubstantiated", "unsupported", "unverified", "unverified", "unverifiable", "unverifiable",]
\end{tcolorbox}
\end{figure*}


\section{Annotation Details}
\label{app:annotation}
\subsection{Annotator Demographics}
\label{app:annotation::demographics}
A total of six annotators were involved in the annotation process. Five are Ph.D.\ students in AI/NLP and one is a full-time NLP research scientist--all fluent speakers of English. None of these individuals received compensation beyond their recognition as co-authors of this work.

\subsection{Annotation Interface}
\label{app:annotation::interface}

\begin{figure*}
    \centering
    \includegraphics[width=\textwidth]{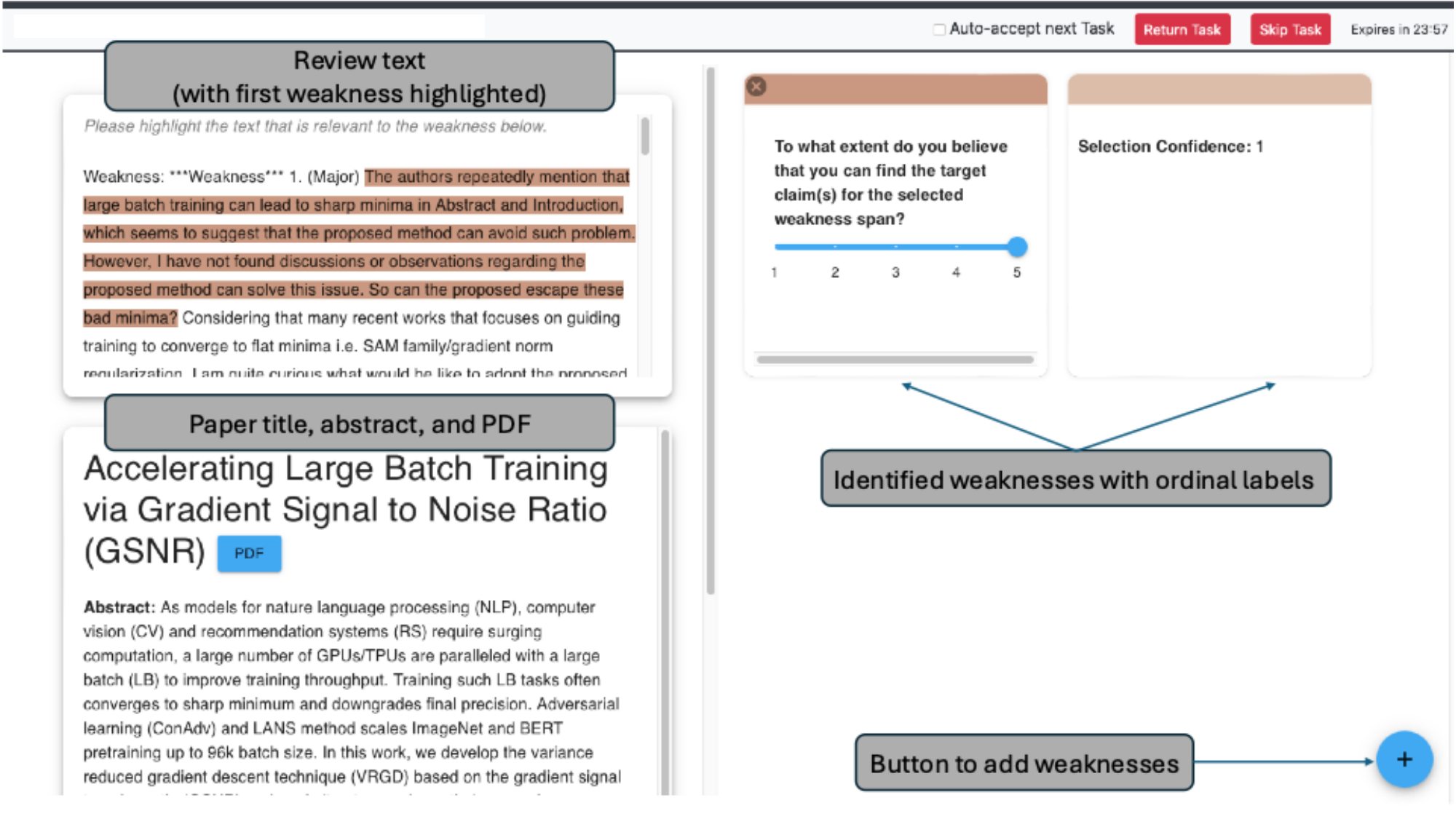}
    \caption{Annotation interface for the \textbf{Weakness Identification (WI)} subtask. Annotators select contiguous spans from from the review text (top left), each describing a weakness raised by the reviewer. For each weakness, annotators supply a Likert-scale judgment (top right) indicating the extent to which they believe the weakness targets a \emph{specific claim} made in the paper (bottom left). Annotators select as many weaknesses as they can find in the review that plausibly target \emph{some} claim. The paper in this example (and in Figures \ref{fig:claim-association-part-1}-\ref{fig:claim-association-part-3}) is \citet{jiang-etal-2023-accelerating}.}
    \label{fig:weakness-identification}
\end{figure*}

\begin{figure*}
    \centering
    \includegraphics[width=\textwidth]{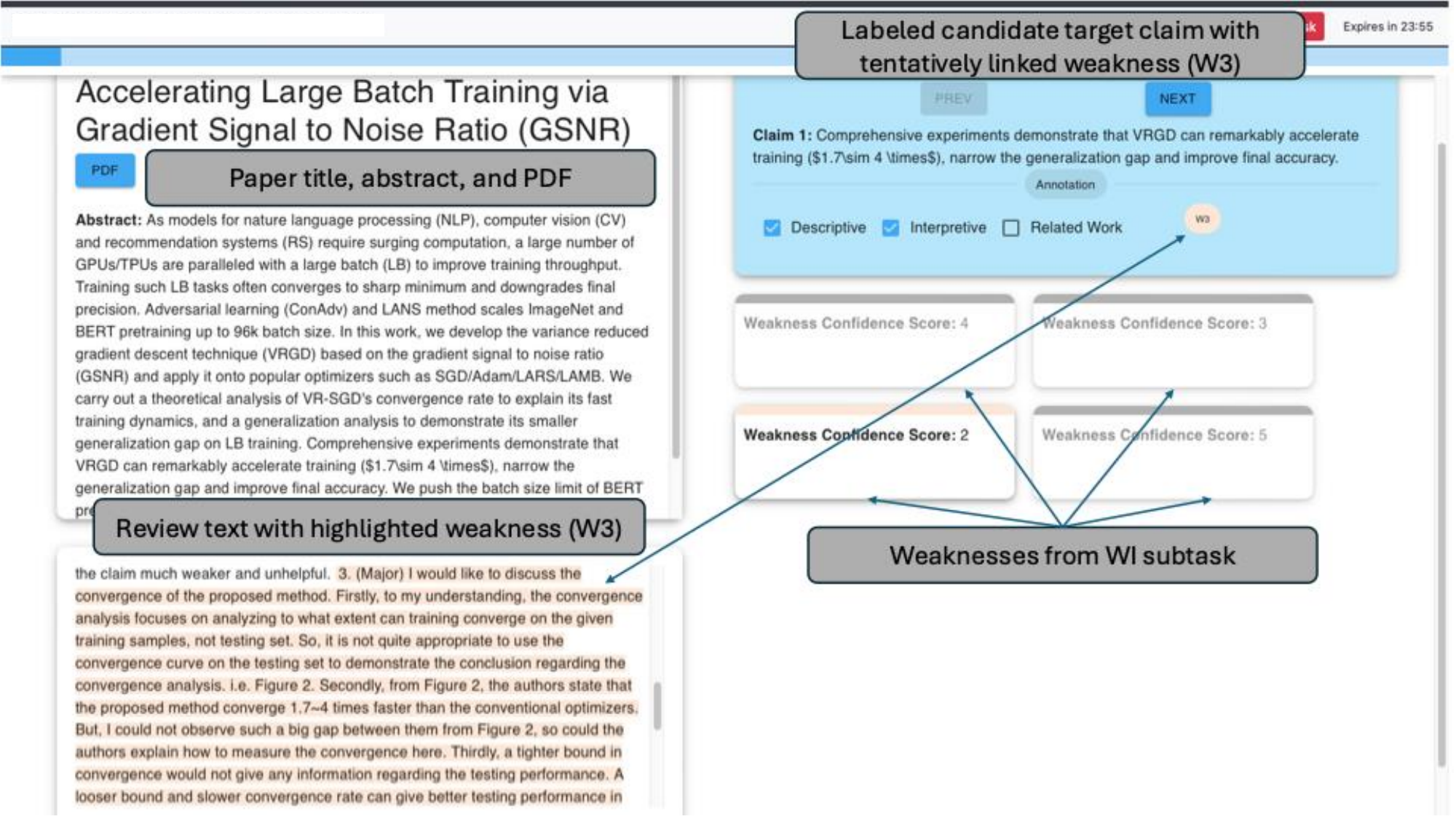}
    \caption{Annotation interface showing part of the \textbf{Claim Association (CA)} subtasks. Given (1) the weaknesses identified for a given review during the Weakness Identification (WI) subtask (\autoref{fig:weakness-identification}) and (2) a set of candidate claims extracted by GPT-4o, annotators must determine which of these claims are targeted by each weakness (if any). Although during the annotation we also ask annotators to provide type labels for each candidate target claim, we find these labels do not provide necessary information for other annotation subtasks or for LLM reasoning and decide to drop it from the final dataset/evaluation.}
    \label{fig:claim-association-part-1}
\end{figure*}

\begin{figure*}
    \centering
    \includegraphics[width=\textwidth]{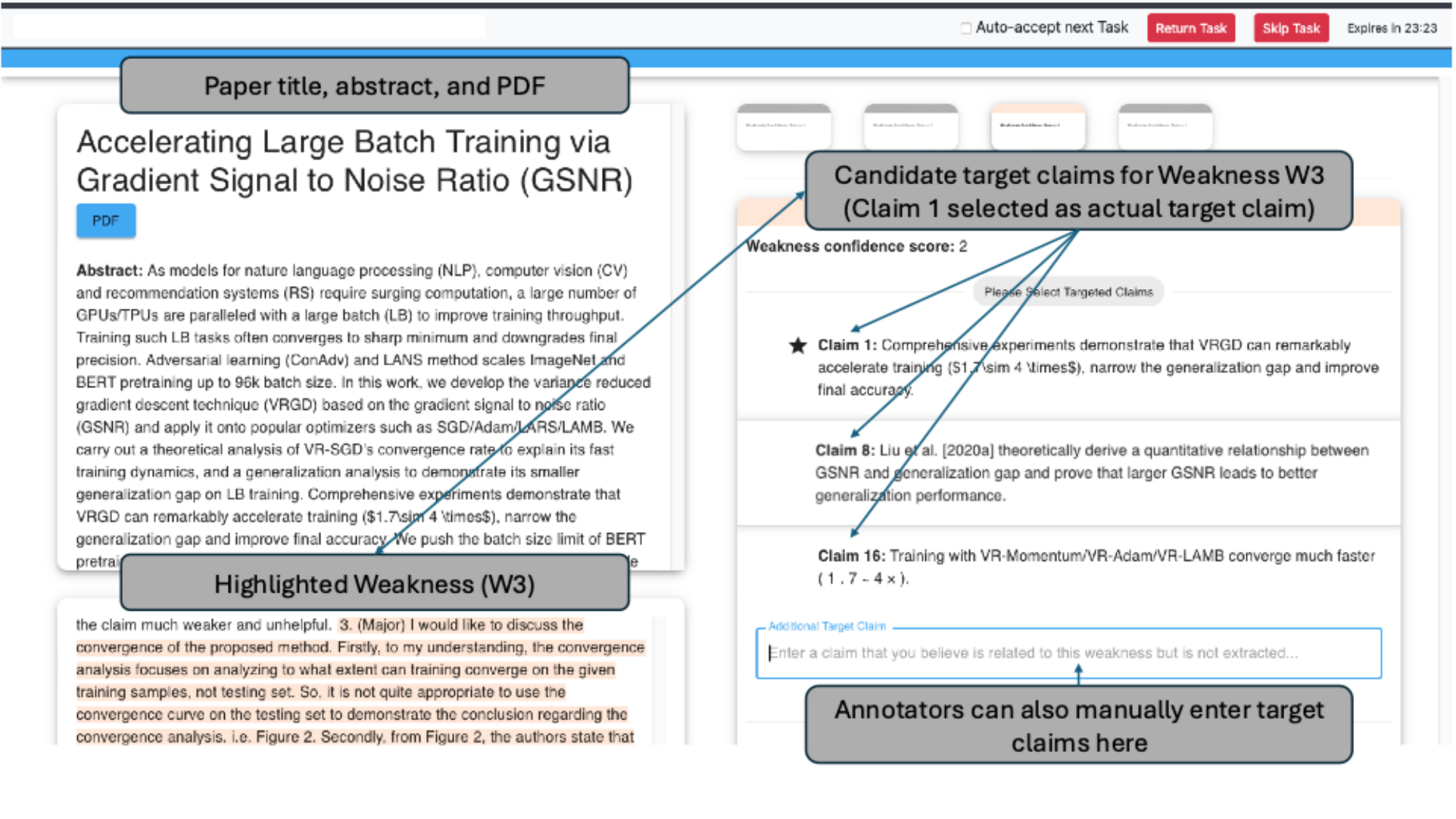}
    \caption{Annotation interface for the final part of the \textbf{Claim Association (CA)} subtask. After selecting a set of \emph{tentative} target claims for each weakness (\autoref{fig:claim-association-part-1}), annotators then \emph{finalize} their selections by starring a (potentially improper) subset of these claims (here, \textbf{Claim 1}). Additionally, they may manually add a target claim from the text if it was not among the extracted candidate claims (bottom right).}
    \label{fig:claim-association-part-2}
\end{figure*}

\begin{figure*}
    \centering
    \includegraphics[width=\textwidth]{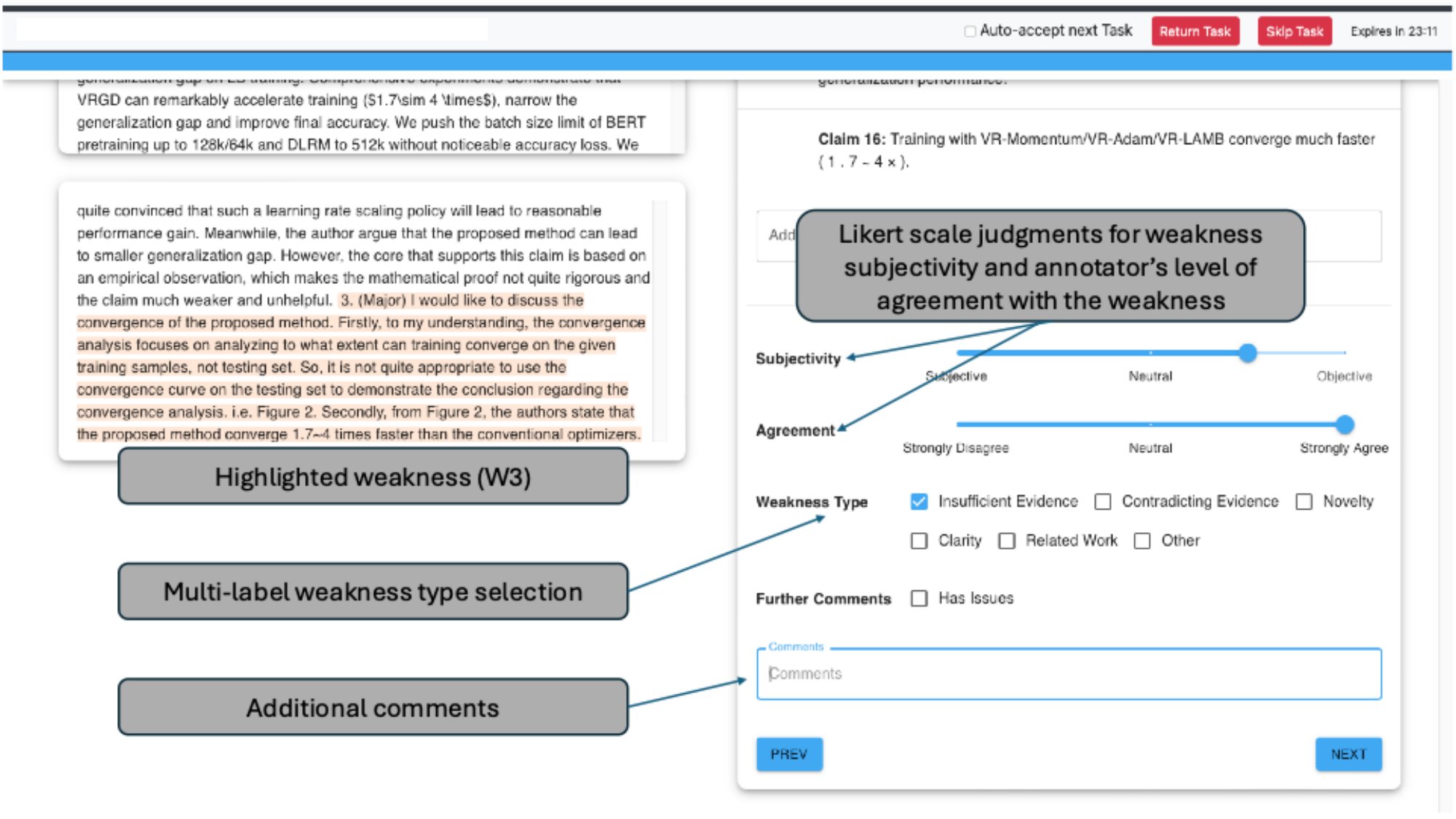}
    \caption{Annotation interface for the \textbf{Weakness Labeling (WL)} subtask. After finalizing the set of target claims for a given weakness (\autoref{fig:claim-association-part-2}), annotators label these weaknesses by providing: (1) a \emph{subjectivity} rating, indicating how subjective the annotator believes the weakness to be; (2) an \emph{agreement} rating, indicating the extent to which the annotator agrees that the weakness is valid; and (3) a multi-label set of \emph{weakness types}, indicating the kind of weakness this is. Annotators may also leave further comments about the weakness in the text box at the bottom.}
    \label{fig:claim-association-part-3}
\end{figure*}

\subsection{Further Annotation Details}
\label{app:annotation::details}
This section provides some additional details about the annotation process. Annotation instructions are included in the supplementary materials.

\paragraph{Weakness Groundedness Labels} Below are descriptions of each value on the ordinal groundedness labeling scale used during the WI annotation subtask.

\begin{enumerate}
    \setcounter{enumi}{-1}
    \item Not an actual scale value (DO NOT USE); included only for reference. This value is reserved for spans of text you aren't even inclined to highlight as potential claim-related weaknesses in the first place. This would include weaknesses that very clearly do not target a claim or result (e.g.\ those that call out poor style or unclear exposition) or other spans that don't describe a weakness at all (e.g.\ spans that summarize related work or that pose a clarifying question).
    \item The weakness \emph{seems} to be responding to some claim or result in the paper (and thus is not a 0), but it's unlikely ($<25$\% chance) you'd be able to find actual claims in the paper that you would consider at all targeted by this weakness. This could be because the weakness is highly subjective or because the reviewer makes lots of inferences not grounded in the paper's contents.
    \item Like (1), but you think it's somewhat likelier ($25$-$50$\% chance) that you'd be able to find at least some claim or result in the paper targeted by this weakness.
    \item The weakness makes reference to a claim that is plausibly grounded in the paper, but that is not an explicit quote or not an obvious paraphrase. You would likely ($50$-$75$\% chance) be able to find a claim or claims targeted by this weakness in the paper, but the actual claims discussed in the weakness might reflect a modest amount of interpretation on the part of the reviewer, and, further, might be made on the basis of figures, tables, or numerical results rather than claims \emph{per se}.
    \item Like (3), but you are \emph{quite} confident ($>75$\% chance) that you would be able to find target claims for this weakness in the paper. The claims referenced in the weakness involve minimal interpretation on the part of the reviewer and are very closely grounded either in claims from the paper and/or in figures, tables, or numerical results.
    \item The weakness explicitly (partially) quotes or otherwise makes explicit reference (e.g.\ via paraphrase) to a specific \emph{claim}---not figure, table, or raw numerical result---that is almost certainly made in the paper (assuming the reviewer is not a blatant liar). These spans may start with (e.g.) ``the paper claims that...'' or ``the authors state that...'', or may refer to specific line numbers that contain the claim of interest.
\end{enumerate}

\paragraph{Weakness Objectivity Labels}
The objectivity score is an ordinal score (1-5) for how objective the criticism raised by a particular weakness is. Below are the interpretations of scores 1, 3, and 5 as given to annotators, where scores of 2 and 4 are to be interpolated on the basis of these descriptions.
\begin{enumerate}
    \item The claim-related weakness depends almost exclusively on subjective judgments about one or more aspects of the paper, such as how significant or exciting its contributions are, its novelty, likely impact, ethical implications, etc.
    \setcounter{enumi}{2}
    \item The claim-related weakness depends on objective observations or judgments but also includes some subjective interpretations of, or opinions about, those observations and their implications.
    \setcounter{enumi}{4}
    \item The claim-related weakness depends almost exclusively on objective observations (possibly in conjunction with valid commonsense, mathematical, logical, or statistical reasoning), with limited or no appeal to subjective interpretation of the paper's claims or contributions.
\end{enumerate}

\paragraph{Weakness Agreement Labels}
the agreement score is an ordinal score (1-5) for a weakness that represents the the extent to which an annotator agrees that the issue raised by the weakness is a problem for the paper. As with the objectivity labels, we provided annotators with descriptions for scores of 1, 3, and 5, with the interpretations of scores of 2 and 4 to be interpolated on the basis of these descriptions.
\begin{enumerate}
    \item The claim-related weakness makes no sense, is ill-founded, or simply does not apply to any claims made in the paper.
    \setcounter{enumi}{2}
    \item The claim-related weakness is somewhat convincing and/or partially applicable to the target claims.
    \setcounter{enumi}{4}
    \item The claim-related weakness is fully convincing and directly applicable to the target claims. The target claims would need to be heavily revised or even jettisoned entirely in response to the weakness.
\end{enumerate}


\paragraph{Weakness Type Labels} Below are the descriptions of the multi-label weakness types as provided to annotators. As with the claim types (see above), our preliminary investigations revealed that a substantial fraction of weaknesses were adequately characterized only by two or more of these labels (e.g.\ weaknesses that call the \emph{novelty} of some method into question based on very similar proposals in uncited \emph{related work}). Thus, we were similarly motivated to implement a multi-label typing scheme here.
\begin{itemize}
    \item \emph{Insufficient Evidence}: The weakness argues that the paper provides insufficient evidence for some claim(s)---e.g.\ due to lack of statistical significance testing, missing experiments, weak baselines, inappropriate choice of datasets, etc.
    \item \emph{Contradictory Evidence}: The weakness provides evidence that some claim(s) in the paper are not only insufficiently supported but are in fact false---e.g.\ due to numerical or methodological errors or results in another paper (see \emph{related work}) that undermine the paper’s claims of state-of-the-art performance.
    \item \emph{Novelty}: The weakness claims that the paper is not novel in one or more important respects.
    \item \emph{Clarity}: The weakness highlights difficulties in understanding the paper itself---possibly due to poor writing or paper organization.
    \item \emph{Related Work}: The weakness calls attention to other work related to the paper that was uncited or otherwise given inadequate consideration or treatment.
    \item \emph{Other}: The weakness identifies some issue with the paper that does not clearly belong to one of the other categories described above.
    \end{itemize}


\section{Experimental Details}
\label{app:experiments}
\subsection{Model Details and Hyperparameters}
\label{app:experiments::model-details}
We run all the experiments (WLE, CA, CV, and reviewer-written weakness grounding for WIS evaluation) with \texttt{GPT-4o-2024-08-06} zero-shot prompting.
We use \texttt{temperature=0.9} for CA and \texttt{temperature=0.3} for all the other experiments.
All the experiments are repeated 3 times with \texttt{seeds=[0,42,2025]} and we report the average results across the three runs.

We provide the prompts for all the experiments in \autoref{app:prompts}

\section{Prompts}
\label{app:prompts}
Prompts used in data preprocessing and experiments.
\begin{figure*}
\begin{tcolorbox}[colback={lightgray},title={\textbf{Claim Extraction}},colbacktitle=white,coltitle=black]
You are an experienced AI and NLP researcher that is going to review a paper. Given the title, abstract, and a chunk of text in the paper, your first task is to extract all the scientific claims the authors make in this chunk. The claims should be a consecutive span of text from the sections and consists of one or more sentences. Make sure to extract the exact original claims from the text, without any paraphrasing. When extracting claims, focus on claims that are with respect to the findings/contributions/results/relation with related work of the research, skip all other claims, especially ignore any descriptions of the ideas, methods, and experiment setup. If the chunk contains no claim satisfies the criteria, simply output an empty list. There might be some noisy text in the chunk, such as ocr text from figures, references, due to the noise in parsing the paper pdf.Ignore and only ignore the noisy text, extract the claims from the rest of the text. You can determine if a part of the chunk is noisy by referring to the title and abstract. Output your results as a JSON object with the following format:
\{Claims: ['Claim 1', 'Claim 2', ...]\}, where the claims are listed in the order they appear in the text.
\end{tcolorbox} 
\end{figure*}
\begin{figure*}
\begin{tcolorbox}[colback={lightgray},title={\textbf{Caption Extraction}},colbacktitle=white,coltitle=black]
Given an image of a table/figure/algorithm from a paper, your task is to extract the caption of the image.The is caption usually located above or below the image, and starts with 'Table X:', 'Figure X:', or 'Algorithm X:', where X is the index of the image.Output your results as a JSON object with the following format: \{"Caption": "The caption of the image"\}
\end{tcolorbox}
\end{figure*}
\begin{figure*}
\begin{tcolorbox}[colback={lightgray},title={\textbf{Text Cleaning}},colbacktitle=white,coltitle=black]
You are an expert in AI/NLP. Given a paragraph extracted from an AI/NLP paper using OCR, your task is to clean the text by removing OCR noises. Specifically, the paper are extracted from NeurIPS2023/2024 anonymized submissions, so OCR will identify the line numbers and embed them in the content text. Additionally, there might be text from tables/figures/captions that are accidentally included in the main text due to OCR error. Your task is to clean these noise strings from the text. Keep the substring such as '’' that represents ''s'. And for all the numbers encoded in brackets, e.g. [20] are in-line citation, only remove them if they are within the span that you determine are wrong extraction from table/figure/captions. Use your knowledge to determine which parts are noise and which parts are original text, based on fluency and coherence. Especially when determining when mentioning tables/figures/captions is intended in the main content or are OCR errors. Do not modify any of the original text, instead, copy them faithfully. Output the cleaned text.Output your results as a JSON object with the following format: \{'cleaned\_text': 'The cleaned text'\}
\end{tcolorbox}
\end{figure*}
\newpage

\begin{figure*}
\begin{tcolorbox}[colback={lightgray},title={\textbf{NLP Topic Classification}},colbacktitle=white,coltitle=black]
You are an experienced AI and NLP researcher that is going to serve as the program chair for a top AI conference. Given a paper title and abstract, and list of keywords, you job is to determine if the paper is broadly relevant to natural language processing (NLP) research. A paper is broadly related to NLP if it's any part of its topic/subject matter/methods/techniques/data and resource use/evaluation is related to any subfield of NLP. Output your results as a JSON object with the following format: 
\{"NLP": "YES/NO"\}, where YES indicates the paper is broadly related to NLP, NO indicates the paper is not related to NLP. .
\end{tcolorbox}
\end{figure*}

\begin{figure*}
\begin{tcolorbox}[colback={lightgray},title={\textbf{Weakness Labeling and Editing (WLE)}},colbacktitle=white,coltitle=black]
You are an experienced AI and NLP researcher that is going to give meta reviews. You are provided with:

1. The full text, as well as its tables, figures and algorithms as images with captions (if any) of the main paper
2. (Optionally) The full text, as well as its tables, figures and algorithms as images with captions of one or more related work
3. A review that comments on some weaknesses of the paper.
4. A span of text extracted from the review that is potentially a **claim-related weakness**.
5. One or more claims from the paper that are **target claim(s)** of the claim-related weakness.
A claim-related weakness is a span of text in the provided review that specifically comments on shortcomings of the paper, usually with reference to particular claims the paper makes. A claim is said to be a target claim of a claim-related weakness if:

1. The weakness clearly disputes the truth or accuracy of that claim.
2. Making this determination does not require appealing to any other claim(s).

Your tasks are to:

1. Give an ***objectivity score*** for the claim-related weakness.
2. Give an ***agreement score*** for the claim-related weakness.
3. Assign one or more ***weakness type label(s)*** to the claim-related weakness.
4. If needed, rewrite the claim-related weakness to make it more sound based on your understanding of (a) the paper and optionally related work, (b) the target claim(s), and (c) the original claim-related weakness.

The objectivity score is an ordinal score (1-5) for the claim-related weakness that represents the degree of objectivity involved in the judgments of the agreement annotation. The interpretations of the values 1, 3, and 5 on this scale are as follows:

1: The claim-related weakness depends almost exclusively on subjective judgments about one or more aspects of the paper, such as how significant or exciting its contributions are, its novelty, likely impact, ethical implications, etc.
3: The claim-related weakness depends on objective observations or judgments but also includes some subjective interpretations of, or opinions about, those observations and their implications.
5: The claim-related weakness depends almost exclusively on objective observations (possibly in conjunction with valid commonsense,mathematical, logical, or statistical reasoning), with limited or no appeal to subjective interpretation of the paper’s claims or contributions.

A score of 2 should be based on an "interpolation" between the descriptions for 1 and 3 above and a score of 4 should be based an "interpolation" between the descriptions for 3 and 5 above.

Next, the agreement score is an ordinal score (1-5) for the claim-related weakness that represents the the extent to which you would agree with its content if you were the meta-reviewer for the paper. The interpretations of the values, 1, 3, and 5 on this scale are as follows:

1: The claim-related weakness makes no sense, is ill-founded, or simply does not apply to any claims made in the paper.
3: The claim-related weakness is somewhat convincing and/or partially applicable to the target claims.
The associated target claims would need to be qualified or rephrased in response to the weakness.
5: The claim-related weakness is fully convincing and directly applicable to the target claims. The target claims would need to be heavily revised or even jettisoned entirely in response to the weakness.

As with the objectivity score, a score of 2 should be based on an "interpolation" between the descriptions for 1 and 3 directly above and a score of 4 should be based an "interpolation" between the descriptions for 3 and 5 directly above.
\end{tcolorbox}
\end{figure*}
\begin{figure*}
\begin{tcolorbox}[colback={lightgray},title={\textbf{Weakness Labeling and Editing (WLE) (Continued)}},colbacktitle=white,coltitle=black]
Finally, the weakness type labels characterize the kind of claim-related weakness we are dealing with. Multiple labels may apply and you must select at least one. The labels are as follows:

- Insufficient Evidence (insufficient): The weakness argues that the paper provides insufficient evidence for some claim(s)—e.g. due to lack of statistical significance testing, missing experiments, weak baselines, inappropriate choice of datasets, etc.
- Contradictory Evidence (contradictory): The weakness provides evidence that some claim(s) in the paper are not only insufficiently supported but are in fact false—e.g. due to numerical or methodological errors or results in another paper that undermine the paper's claims of state-of-the-art performance.
- Novelty (novelty): The weakness claims that the paper is not novel in one or more important respects.
- Clarity (clarity): The weakness highlights difficulties in understanding the paper itself—possibly due to poor writing or paper organization.
- (Missing) Related Work (related\_work): The weakness calls attention to other work related to the paper that was uncited or otherwise given inadequate consideration or treatment.
- Other (other): The weakness identifies some issue with the paper that does not clearly belong to one of the other categories described above.

Your output must be a JSON object with the following format: \{"Reasoning Objectivity": "Your reasoning for the objectivity score", "Objectivity Score": "The objectivity score", "Reasoning Agreement": "Your reasoning for the agreement score", "Agreement Score": "The agreement score", "Reasoning Weakness Type": "Your reasoning for the weakness type label(s)", "Weakness Types": \{"insufficient": true/false, "contradictory": true/false, "novelty": true/false, "clarity": true/false, "related\_work": true/false, "other": true/false\}\}"Reasoning Rewritten Weakness": "Your reasoning for if the claim-related weakness span needs to be rewritten and how", "Rewritten Weakness": "The claim-related rewritten weakness span"\}
\end{tcolorbox}
\end{figure*}
\newpage
\begin{figure*}
\begin{tcolorbox}[colback={lightgray},title={\textbf{Claim Association (CA)}},colbacktitle=white,coltitle=black]
You are an experienced AI and NLP researcher that is going to write meta-reviews. You are provided with: 

1. The full paper text and a numbered list of claims that have been extracted from the paper.
2. A review that comments on some weaknesses of the paper.
3. A span of text extracted from the review that is potentially a **claim-related weakness**. A claim-related weakness is a span of text in the above review that specifically comments on shortcomings of the paper, usually with reference to particular claims the paper makes.
4. A weakness confidence score: An ordinal label (1-5) indicating how likely you think it is that the claim-related weakness has at least one **target claim** in the paper.

Your tasks is to :
 Select a subset of claims from the provided claim list that are **target claims** of the claim-related weakness.
 A claim is said to be a target claim of a claim-related weakness if:

1. The weakness clearly disputes the truth or accuracy of the claim.
2. Making this determination does not require appealing to any other claim(s).

Concerning point (2), a weakness, if true, can clearly have implications for the truth or accuracy of multiple claims made by a paper. But for our purposes, we want to focus only on the claims that are most immediately disputed, which is why we stipulate (2) above. We might therefore distinguish ***direct target*** claims from ***indirect target*** claims—claims whose truth or accuracy is affected by some weakness (if true), but only in virtue of other claims. We illustrate this distinction with the example below.

Example 1:
----------
Weakness 1: The paper’s claim that method X demonstrates superior performance over all baselines is not convincing since the confidence intervals of X’s performances largely overlap with many of the baselines’ confidence intervals.
Claim 1: The results in Table 2 demonstrate the superior performance of proposed method X over all the existing baselines on dataset A.
Claim 2: Findings from Table 1, 2, and 3 showcase the effectiveness of the proposed method X on task T.

Explanation: Here, Claim 1 is a direct target of Weakness 1, since Claim 1’s veracity is directly disputed by Weakness 1, and one need not appeal to any other claims to see that this is so. In contrast, Claim 2 is an indirect target of Weakness 1, since Weakness 1 undermines Claim 2, but only by virtue of Claim 1. You should therefore annotate only Claim 1 as a (direct) target claim.

Another important distinction in target claim association annotation is the one between ***direct target*** claims and merely ***relevant*** claims. You should ***NOT*** associate claims that are merely relevant to some weakness. The following example illustrates this second distinction.

Example 2:
----------
Weakness 2: While the paper claims the introduced module Y enhances the robustness of model M under realistic types of noise, the only datasets that the paper experiments on—i.e. B and C—are either synthetic or make heavily simplifying assumptions about the noise distribution. More realistic datasets like D should also be considered.
Claim 3: Experimental results demonstrate the effectiveness of proposed module Y that renders model M more robust against realistic noise.
Claim 4: As shown in Figure 3 and 4, adding Y to M helps improve the robustness of M under various kinds of noise presented in dataset B and C.

Explanation: Here, Claim 3 is clearly a direct target of Weakness 2. But Claim 4, although topically relevant to Weakness 2, is not a direct target. Even though it refers to datasets B and C, which are mentioned in Weakness 2, Claim 4 is not undermined by Weakness 2 and therefore should not be associated with it.
\end{tcolorbox}
\end{figure*}
\begin{figure*}
\begin{tcolorbox}[colback={lightgray},title={\textbf{Claim Association (CA) (Continued)}},colbacktitle=white,coltitle=black]
For cases where a weakness quotes or mentions a particular claim (principally, weaknesses with a label of 5), the target claim will generally be quite easy to identify. Beyond this, target claims can be trickier to identify, but here are some general principles:

- Take your cue from what the weakness is about. If the weakness is about novelty, an appropriate target claim really ought to be one that makes some assertion about, or else strongly implies, novelty. Or if the weakness is about the superiority of a proposed method relative to existing methods, you ought to be able to find a claim to that effect (or one that strongly implies that superiority) in the paper—not just a table with results. This is a fairly basic point, but the moral is that if the paper doesn't actually make the claim imputed to it by the weakness, then that weakness might just not have a target claim. Don't go scrounging for target claims that aren't there.
- Relatedly, if the weakness is very broad or vague (typically, these will have a label of 1 or 2), then they probably don't have a target claim either.
- However, if you think the claim-related weakness should have a target claim but you cannot find one in the list of claims, you may copy up to one additional target claim from the paper text (called a **custom target claim**). You should always use this option if a weakness quotes or mentions a claim in the paper that does not appear anywhere in the list of candidate claims.
- Additionally, even if the weakness does *not* explicitly quote or mention a specific claim, you may still be able to find a target claim in the paper. You should use a ***custom target claim*** in this situation as well—especially if the ordinal label score is relatively high (3-5) for the weakness but you are struggling with finding a proper target claim in the list of claims.
You should select ***AT MOST 3*** target claims for the weakness, including the custom claim (if you use one).

Your output must be a JSON object with the following format: \{"Reasoning": "Your reasoning about why the selected claim(s) are the target(s) of the weakness span, and their labels" "Target claims extracted": ["Claim X: ...", "Claim Y: ..."] (the target claim extracted from the list of claims, if any. Copy the original claim text and the claim number. Leave empty if no target claim is identified.), "Custom target claim": "The custom target claim you extract (if you extract one)\}", 
\end{tcolorbox}
\end{figure*}
\newpage
\begin{figure*}
\begin{tcolorbox}[colback={lightgray},title={\textbf{Claim Verification (CV)}},colbacktitle=white,coltitle=black]
You are an experienced AI and NLP researcher that is going to give meta reviews. You are provided with:

1. The full text, as well as its tables, figures and algorithms as images with captions (if any) of the main paper
2. (Optionally) The full text, as well as its tables, figures, and algorithms as images with captions (if any)  of one or more related work
3. One target claim from the paper.

Your task is to:
Determine whether the target claim exhibits one or more of the following types of weakness: 

- Insufficient Evidence (insufficient): the paper provides insufficient evidence for the target claim, e.g. due to lack of statistical significance tests, missing experiments, weak baselines, inappropriate choice of datasets, etc.
- Contradictory Evidence (contradictory): the target claim in the paper is not only insufficiently supported but is in fact false, e.g. due to numerical or methodological errors or results in another paper.
- Novelty (novelty): novelty asserted in the target claim is not valid in one or more important respects.
- Clarity (clarity): the claim is difficult to understand, possibly due to poor writing or paper organization.
- Related Work (related\_work): the claim fails to take into account critical prior work related to the claim.
- Other (other): there are some other weakness(es) in the target claim that are not covered by any of the above categories.

The target claim definitely exhibits AT LEAST ONE of these types of weakness and may exhibit multiple.You will then need to extract all relevant pieces of evidence from the paper (and related work if any), which may include statements in the text, figures, tables, or algorithms. You must assess ONLY the target claim. DO NOT try to asses any other claims in the paper for weaknesses. This means that when you extract evidence, you must focus ONLY on pieces of evidence that are relevant to the target claim. You must also label each piece of evidence you extract as follows:

- Label each piece of textual evidence as <text\_0>, < text\_1>, ..., based on their order of occurrence in the paper. - Label each piece of figure/table/algorithm evidence as <figure\_x>, <table\_y>, <algorithm\_z> ...,  where x, y and z are the indices of the figures/tables/algorithms as given in their captions.

In your output, you must also explain your REASONING for the types of weakness you think the target claim exhibits. When explaining your reasoning, you should explicitly cite relevant pieces of evidence that you extracted. For example: 'Based on ... in <text\_0> and ... in <figure\_1> and ... in <algorithm\_3>, ... the target claim exhibits...' Please output your results as a JSON object with the following format:

\{"Main Paper Evidence": \{"Text Evidence": \{"text\_0": "The piece of evidence text", "text\_1": "The piece of evidence text", ...\}, "Figure Evidence": ["figure\_x", ..., ], "Table Evidence": ["table\_y", ..., ], "Algorithm Evidence": ["algorithm\_z", ..., ]\}, "Related Work 1 Evidence": \{same as above, if provided and needed \}"Related Work 2 Evidence": \{same as above, if provided and needed \}"Reasoning": "Your reasoning about the weaknesses exhibited by the target claim, specifically stating what part(s) of the target claim exhibit weakness(es) and why", "Weaknesses": \{"insufficient": true/false, "contradictory": true/false, "novelty": true/false, "clarity": true/false, "related\_work": true/false, "other": true/false\}\}
\end{tcolorbox}
\end{figure*}
\newpage
\begin{figure*}
\begin{tcolorbox}[colback={lightgray},title={\textbf{Claim Verification (CV) Distill Reviewer-Written Claim-Related Weakness}},colbacktitle=white,coltitle=black]
You are an experienced AI and NLP researcher that is going to give meta reviews. You are provided with:

1. The full text, as well as its tables, figures and algorithms as images with captions (if any) of the main paper
2. (Optionally) The full text, as well as its tables, figures, and algorithms as images with captions of one or more related work
3. A review that comments on some weaknesses of the paper.
4. A span of text extracted from the review that is potentially a **claim-related weakness**.
5. The types for the claim-related weakness listed above, which might be one or more of the following:
- Insufficient Evidence (insufficient): The weakness argues that the paper provides insufficient evidence for some claim(s)—e.g. due to lack of statistical significance testing, missing experiments, weak baselines, inappropriate choice of datasets, etc.
- Contradictory Evidence (contradictory): The weakness provides evidence that some claim(s) in the paper are not only insufficiently supported but are in fact false—e.g. due to numerical or methodological errors or results in another paper that undermine the paper's claims of state-of-the-art performance.
- Novelty (novelty): The weakness claims that the paper is not novel in one or more important respects.
- Clarity (clarity): The weakness highlights difficulties in understanding the paper itself—possibly due to poor writing or paper organization.
- (Missing) Related Work (related\_work): The weakness calls attention to other work related to the paper that was uncited or otherwise given inadequate consideration or treatment.
- Other (other): there are some other weakness(es) in the target claim that are not covered by any of the above categories.
6. One claim from the paper that is the **target claim** of the claim-related weakness.

A claim-related weakness is a span of text in the provided review that specifically comments on shortcomings of the paper, usually with reference to particular claims the paper makes. A claim is said to be a target claim of a claim-related weakness if:

1. The weakness clearly disputes the truth or accuracy of that claim.
2. Making this determination does not require appealing to any other claim(s).

Your task is to:

Provide the underlying REASONING of the claim-related weakness and its type by grounding it to pieces of evidence from the main and related work (if any), Specifically, explain your REASONING for the types of weakness you think the target claim exhibits, based on the claim-related weakness.based on your understanding of the main paper and optionally related work, (b) the target claim, and (c) the original claim-related weakness and the types of weaknesses the claim-related weakness exhibits. You will first need to extract all relevant pieces of evidence from the paper (and related work if any), which may include statements in the text, figures, tables, or algorithms. You must center your REASONING ONLY around the (target claim, claim-related weakness) pair. DO NOT try to asses any other claims in the paper for weaknesses, or try to justify any other parts of the review. This means that when you extract evidence, you must focus ONLY on pieces of evidence that are relevant to the (target claim, claim-related weakness) pair. You can and should elaborate more in your reasoning through grounding the claim-related weakness in evidence, especially if the original claim-related weakness is too broad with detailed reasoning omitted. In the meantime you must try your best to reflect the original meaning conveyed by the claim-related weakness. But you must try not to directly quote or copy the claim-related weakness, essentially, the REASONING should be a standalone justification of the claim-related weakness. 
\end{tcolorbox}
\end{figure*}
\begin{figure*}
\begin{tcolorbox}[colback={lightgray},title={\textbf{Claim Verification (CV) Distill Reviewer-Written Claim-Related Weakness (Continued)}},colbacktitle=white,coltitle=black]
You must also label each piece of evidence you extract as follows:

- Label each piece of textual evidence as <text\_0>, < text\_1>, ..., based on their order of occurrence in the paper. - Label each piece of figure/table/algorithm evidence as <figure\_x>, <table\_y>, <algorithm\_z> ...,  where x, y and z are the indices of the figures/tables/algorithms as given in their captions.

When explaining your reasoning, you should explicitly cite relevant pieces of evidence that you extracted. For example: 'Based on ... in <text\_0> and ... in <figure\_1>, ... the target claim exhibits...' Please output your results as a JSON object with the following format:

\{"Main Paper Evidence": \{"Text Evidence": \{"text\_0": "The piece of evidence text", "text\_1": "The piece of evidence text", ...\}, "Figure Evidence": ["figure\_x", ..., ], "Table Evidence": ["table\_y", ..., ], \}"Algorithm Evidence": ["algorithm\_z", ..., ]\}, "Related Work 1 Evidence": \{same as above, if provided and needed \}"Related Work 2 Evidence": \{same as above, if provided and needed \}"Reasoning": "Your reasoning about the weaknesses exhibited by the target claim based on the claim-related weakness provided, specifically stating what part(s) of the target claim exhibit weakness(es) and why".
\end{tcolorbox}
\end{figure*}

\section{Reviewer Guideline Examples}
We present in \autoref{tab:reviewer-guideline} several examples of advice for claim-centric criticism from top-tier AI/NLP conferences.
\begin{table*}[t]
\centering
\small
\begin{tabular}{ll}\toprule
                    & \textbf{Reviewer guideline excerpts advising \emph{claim-centric} criticism}\\\hline
\textbf{NeurIPS}             & \begin{tabular}[c]{@{}l@{}}Quality: Is the submission technically sound? Are claims well supported (e.g., by theoretical analysis  \\or experimental results)?
\end{tabular}\\\hline
\textbf{ICLR}                & \begin{tabular}[c]{@{}l@{}}Does the paper support the claims? This includes determining if results, whether theoretical or \\empirical, are correct and if they are scientifically rigorous. \end{tabular}\\\hline
\begin{tabular}[c]{@{}l@{}}\textbf{ARR} \\ \end{tabular}\ & \begin{tabular}[c]{@{}l@{}}Inappropriate scope of the claims: The authors evaluate a sample that does not represent \\the population about which the claim is made. \\
Hypotheses/speculations presented as conclusions:  Every claim that is made has to be based on \\evidence or arguments (the authors' or from other work), or clearly marked as conjecture/speculation.\\ Misleading or inappropriate framing, overclaiming:  E.g., concluding from benchmark evaluation \\ that LLMs generally 'understand' language, without validating that construct\end{tabular}\\\bottomrule
\end{tabular}
\caption{Excerpts from reviewer guidelines of top AI/ML/NLP venues that advise specific, claim-centric reviews.
}
\label{tab:reviewer-guideline}
\end{table*}

\end{document}